\documentclass[11pt]{article}

\usepackage[preprint]{acl}

\usepackage{times}
\usepackage{latexsym}
\usepackage[T1]{fontenc}
\usepackage{amssymb}
\usepackage{booktabs}
\usepackage{makecell}
\usepackage[table]{xcolor}
\usepackage{enumitem}
\usepackage[most]{tcolorbox}

\usepackage{algorithm}
\usepackage{algpseudocode}
\usepackage{booktabs}
\usepackage{tabularx}
\usepackage{array}
\usepackage{amssymb} 
\usepackage{subcaption}

\usepackage[most]{tcolorbox}
\usepackage{enumitem}
\usepackage{xcolor}
\usepackage{hyperref}
\usepackage[utf8]{inputenc}

\tcbuselibrary{breakable, skins}
\usepackage{listings}
\usepackage{microtype}

\usepackage{inconsolata}
\usepackage{amsmath}
\usepackage{graphicx}

\title{WIST: Web-Grounded Iterative Self-Play Tree for Domain-Targeted Reasoning Improvement}

\author{
	Fangyuan Li\textsuperscript{1,2},
	Pengfei Li\textsuperscript{1,2},
	Shijie Wang\textsuperscript{3},
	Junqi Gao\textsuperscript{1},
	Jianxing Liu\textsuperscript{1,$^\dag$},
	Biqing Qi\textsuperscript{3,$^\dag$},
	Yuqiang Li\textsuperscript{2,$^\dag$}
	\\
	\textsuperscript{1}Harbin Institute of Technology \\
	\textsuperscript{2}Shanghai Innovation Institute \\
	\textsuperscript{3}Shanghai Artificial Intelligence Laboratory \\
    {\tt\small lifangyuan@stu.hit.edu.cn, jx.liu@hit.edu.cn}, \\
    {\tt\small \{lipengfei0208, gjunqi97, qibiqing7\}@gmail.com, } \\
    {\tt\small shijie.wang2022@outlook.com, liyuqiang@pjlab.org.cn,}
}

\makeatletter
\newcommand{\blfootnote}[1]{%
	\begingroup
	\renewcommand\thefootnote{}\footnote{#1}%
	\addtocounter{footnote}{-1}%
	\endgroup
}
\makeatother

\begin{document}
\maketitle
\blfootnote{\textsuperscript{$^\dag$}Corresponding Authors}

\begin{abstract}
Recent progress in reinforcement learning with verifiable rewards (RLVR) offers a practical path to self-improvement of language models, but existing methods face a key trade-off: endogenous self-play can drift over iterations, while corpus-grounded approaches rely on curated data environments. We present \textbf{WIST}, a \textbf{W}eb-grounded \textbf{I}terative \textbf{S}elf-play \textbf{T}ree framework for domain-targeted reasoning improvement that learns directly from the open web without requiring any pre-arranged domain corpus. WIST incrementally expands a domain tree for exploration, and retrieves and cleans path-consistent web corpus to construct a controllable training environment. It then performs Challenger--Solver self-play with verifiable rewards, and feeds learnability signals back to update node posteriors and guide subsequent exploration through an adaptive curriculum. Across four backbones, WIST consistently improves over the base models and typically outperforms both purely endogenous self-evolution and corpus-grounded self-play baselines, with the Overall gains reaching \textbf{+9.8} (\textit{Qwen3-4B-Base}) and \textbf{+9.7} (\textit{OctoThinker-8B}). WIST is also domain-steerable, improving \textit{Qwen3-8B-Base} by \textbf{+14.79} in medicine and \textit{Qwen3-4B-Base} by \textbf{+5.28} on PhyBench. Ablations further confirm the importance of WIST's key components for stable open-web learning. Our Code is available at  \url{https://github.com/lfy-123/WIST}.

\end{abstract}
\section{Introduction}
Self-improvement of large language models (LLMs) without human supervision~\cite{clune2019ai,pourcel2025self} is a key step toward more general intelligence. Recent progress in reinforcement learning with verifiable rewards (RLVR)~\cite{hurst2024gpt,guo2025deepseek} shows that when feedback comes from automatically checkable outcomes (e.g., mathematical correctness, program execution, or deterministically verifiable structured outputs), LLMs' reasoning can be reliably strengthened. Unlike costly human annotation, such feedback can be generated at scale, enabling iterative generate--evaluate--update cycles at low marginal cost and offering a practical path toward self-evolving language models.

\begin{figure}[t]
  \includegraphics[width=\columnwidth]{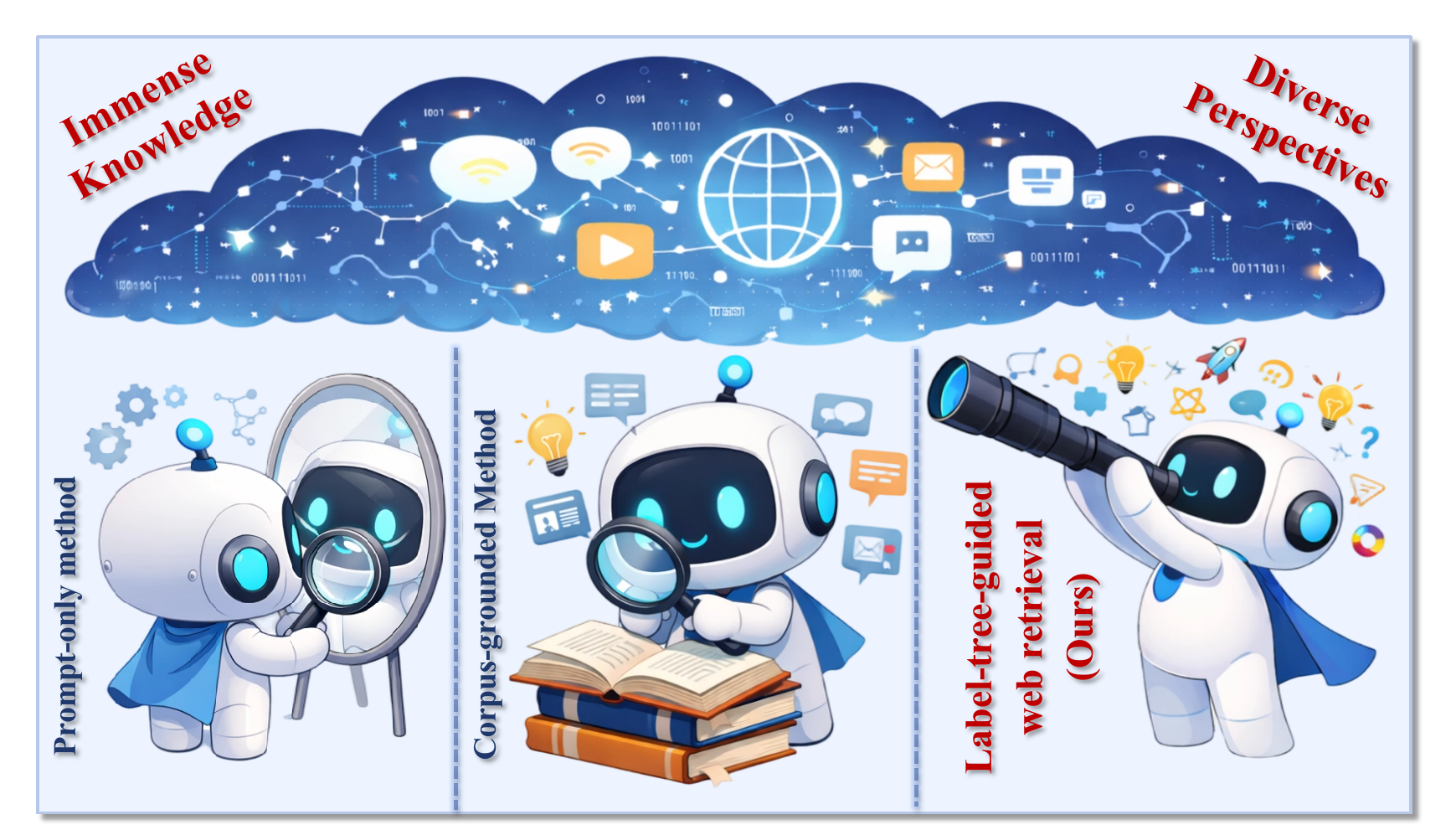}
  \caption{\textbf{Comparison of R-Zero, SPICE, and our WIST.}}
  \label{fig:rz_spice_ours}
\end{figure}

Motivated by this goal, prior work has explored several routes to self-improvement. A common approach bootstraps from seed data or existing task collections via self-training and synthetic data generation (e.g., STaR~\cite{zelikman2022star}, MetaMath~\cite{yu2023metamath}, Self-Instruct~\cite{wang2023self}). Another line~\cite{zhao2025absolute} adopts self-play and automatic curricula in verifiable environments (especially code), often implementing \emph{generate--verify--learn} loops with adversarial or cooperative role specialization. More recently, R-Zero~\cite{huang2025r} pursues fully endogenous self-evolution by creating tasks from zero external data and deriving rewards from internal signals such as self-consistency. In contrast, SPICE~\cite{liu2025spice} emphasizes external knowledge: it treats a large corpus as an environment and uses corpus-grounded verifiable QA under information asymmetry to mitigate hallucination accumulation and stagnation. Despite these advances, a core tension remains: purely endogenous generation can drift and degrade over iterations, while corpus-dependent approaches rely on curated sources and often struggle to cover specialized domains, shifting the burden to building and maintaining high-quality data environments.

We propose \textbf{WIST}, a \textbf{W}eb-grounded \textbf{I}terative \textbf{S}elf-play \textbf{T}ree framework that improves reasoning by enabling models to discover and learn domain-relevant knowledge from the open web. The open web is rich but noisy and unstructured, making domain-relevant verifiable signals hard to extract. Inspired by prior work on structuring such data for learning~\cite{gao2025bohdi,cao2025condor}, WIST organizes exploration with a dynamically expanding domain tree: starting from a user-specified domain label, the model incrementally decomposes the domain into finer-grained concepts down to leaf-level knowledge points. Each sampled root-to-leaf path then triggers corpus acquisition, where WIST retrieves and cleans path-consistent web documents to construct a lightweight, continuously refreshed corpus pool. Conditioned on the retrieved corpus, WIST runs a Challenger--Solver self-play loop with verifiable rewards, and converts the resulting learnability feedback into node-wise posterior updates. These posteriors guide subsequent path sampling, yielding an adaptive curriculum that increasingly focuses on the model's weak yet learnable regions. Compared with fully endogenous self-evolution (e.g., R-Zero), WIST grounds training in retrieved corpus to mitigate signal drift; compared with corpus-grounded self-play (e.g., SPICE), WIST removes reliance on a fixed curated corpus by expanding coverage through structured open-web exploration (Figure~\ref{fig:rz_spice_ours}). We present additional related work in Appendix~\ref{sec:related}.

Empirically, WIST delivers consistent gains across diverse backbones. For example, on \textit{Qwen3-4B-Base}, WIST improves the Overall score from 33.3 to 43.1 (+9.8), outperforming R-Zero (40.6) and SPICE (41.8); on \textit{Qwen3-8B-Base}, it reaches 46.7 (vs.\ 42.1 base), exceeding R-Zero (45.5) and SPICE (46.0); and on \textit{OctoThinker-8B-Hybrid-Base}, it improves from 22.9 to 32.6 (+9.7), surpassing both baselines. Moreover, WIST is inherently domain-steerable: by switching only the target domain label to \textbf{physics}, it yields measurable gains on PhyBench (EED score 4.73 $\rightarrow$ 10.01 in 50 steps), demonstrating that open-web corpus can support domain-specific self-evolution without relying on any carefully curated domain corpus. Ablations further show that reward-guided exploration stabilizes training and that the tree structure is essential for maintaining coverage and reliably mining high-value knowledge from the open web.

Our contributions include:
\begin{itemize}[nosep]
  \item We introduce \textbf{WIST}, a web-grounded self-play Tree framework that enables \emph{domain-targeted} reasoning improvement without requiring a manually curated domain corpus.
  \item We propose a dynamically expanding domain tree with posterior-guided path sampling, which structures open-web exploration and induces an adaptive curriculum from learnability feedback.
  \item We demonstrate strong gains on mathematical and general reasoning benchmarks across multiple backbones, and validate domain steering to physics through systematic ablations.
\end{itemize}

\begin{figure*}[t]
  \centering
  \includegraphics[width=0.93\textwidth]{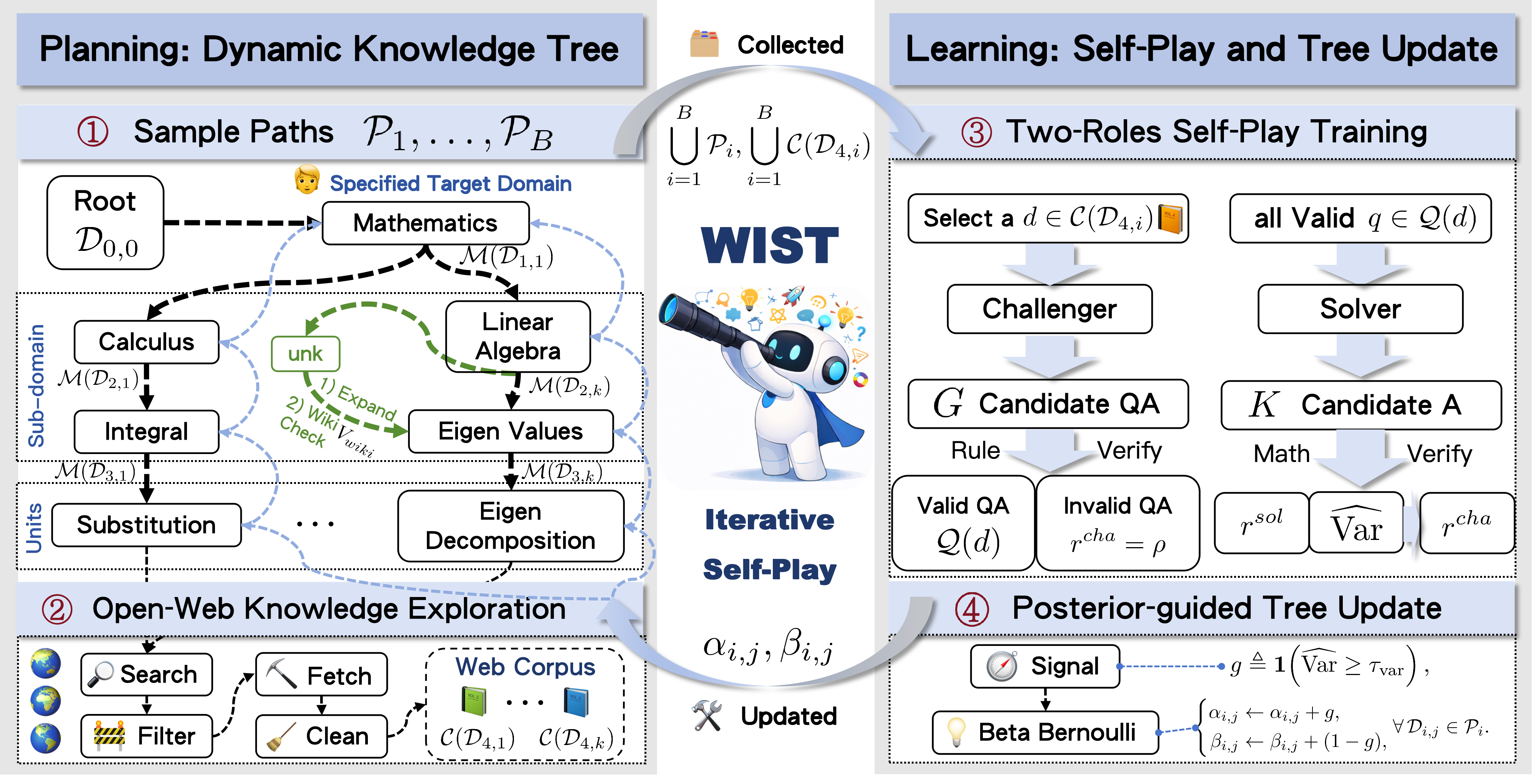}
  \caption{Overview of our proposed WIST, a web-grounded iterative self-play Tree framework for domain-targeted reasoning improvement.}
  \label{fig:overview}
\end{figure*}

\section{Preliminaries}
\subsection{Reinforcement Learning with Verifiable Rewards}
Reinforcement Learning with Verifiable Rewards (RLVR) is a paradigm for fine-tuning models in domains where response quality can be deterministically verified. Given a prompt (question) $x$, a policy LLM $\pi_{\theta}$ generates an answer $\hat{y}\sim\pi_{\theta}(\cdot\mid x)$. RLVR assumes a rule-based verifier
\begin{equation}
v:\ \mathcal{Y}\times \mathcal{Y}\rightarrow\{0,1\},
\end{equation}
which compares a generated answer $\hat{y}$ against a reference answer $y^*$ and returns $1$ if $\hat{y}$ is equivalent to $y^*$ under task-specific criteria
(e.g., normalized exact match, symbolic equivalence, or deterministic format constraints), and $0$ otherwise.
This induces a binary reward:
\begin{equation}
r(\hat{y};y^*) \triangleq v(\hat{y},y^*).
\end{equation}
Such verifiable rewards are especially effective for tasks with unambiguous correctness (e.g., mathematical reasoning) and form the basis of the Solver training reward in our work.

\subsection{Group Relative Policy Optimization Done Right}
We optimize $\pi_\theta$ using Dr.\ GRPO~\cite{liu2025understanding}, a group-based policy optimization method tailored to RLVR that avoids value-function fitting.
For each prompt $x$, we sample a group of $G$ responses $\{\hat{y}_i\}_{i=1}^{G}$ and compute rewards $\{r_i\}_{i=1}^{G}$.
Dr.\ GRPO uses a group-centered advantage:
\begin{equation}
A_i \triangleq r_i - \frac{1}{G}\sum_{j=1}^{G} r_{j},
\end{equation}
and applies a PPO-style clipped objective at the token level with a \emph{global} normalization constant (instead of length-normalization), reducing length-related optimization bias. In our implementation, Dr.\ GRPO is used as the advantage computation for our training pipeline.

\section{Methodology}
\subsection{Overview}

We propose \textbf{WIST}, a web-grounded iterative self-play tree framework for domain-targeted reasoning improvement (Algorithm in Appendix~\ref{sec:algorithm}; overview in Figure~\ref{fig:overview}). WIST closes the loop between \emph{where to explore} and \emph{what is learned} by sampling a path on a dynamically expanding domain tree, retrieving and cleaning path-aligned web documents into leaf-level corpus pools, running corpus-grounded Challenger--Solver self-play with verifiable rewards, and feeding the resulting learnability signal back to update node posteriors for subsequent exploration and curriculum budget allocation.

WIST has three coupled components: (1) a \textbf{self-expanding domain tree} for exploration planning that decomposes the target domain into leaf-level concepts; (2) an \textbf{open-web retrieval, filtering, and corpus construction pipeline} that constructs and attaches cleaned corpus pools to leaf nodes; and (3) a \textbf{tree-guided curriculum} that updates node-wise Beta posteriors and performs Thompson sampling with a sliding-window update to track non-stationary learning.

\subsection{Dynamic Domain Tree Construction}
To support controllable and scalable exploration in an open-web environment, we organize the target domain as a hierarchical tree and expand it incrementally during training. The tree decomposes coarse-grained topics into finer-grained concepts, so that sampled leaf nodes correspond to searchable and verifiable knowledge units for subsequent web retrieval and self-play.

\paragraph{Hierarchical domain tree $\mathcal{T}$.}

We represent the target domain as a directed tree $\mathcal{T}=(\mathcal{V},\mathcal{E})$ with maximum depth $\mathcal{L}$. Nodes are organized into layers $\{\mathcal{V}_i\}_{i=0}^{\mathcal{L}}$, where $\mathcal{V}=\bigcup_{i=0}^{\mathcal{L}}\mathcal{V}_i$. We define a virtual root node $\mathcal{D}_{0,0} \in \mathcal{V}_0$, whose child nodes correspond to targeted domain (e.g., mathematics, physics). To support continual expansion, each layer contains discovered topic nodes and an additional unknown placeholder node:

\begin{equation}
\mathcal{V}_i=\{\mathcal{D}_{i,1},\dots,\mathcal{D}_{i,N_i}\}\ \cup\ \{\mathcal{D}_{i,\mathrm{unk}}\}.
\end{equation}
The edge set encodes parent--child relations across layers:

\begin{equation}
\resizebox{0.85\linewidth}{!}{$
\begin{aligned}
\mathcal{E}=\{(\mathcal{D}_{i,j},\mathcal{D}_{i+1,k}) \mid\;& \mathcal{D}_{i+1,k}\ \text{is a child of}\ \mathcal{D}_{i,j},\\
& i\in  \{0,\dots,\mathcal{L}-1\}\}.
\end{aligned}
$}
\end{equation}

A root-to-leaf path is
\begin{equation}
\mathcal{P}=[\mathcal{D}_{0,0}\rightarrow \mathcal{D}_{1,i_1}\rightarrow \cdots \rightarrow \mathcal{D}_{\mathcal{L},i_\mathcal{L}}],
\end{equation}
where the leaf node $\mathcal{D}_{\mathcal{L},i_\mathcal{L}}$ is treated as a minimal knowledge unit that triggers subsequent web retrieval and corpus sampling.

\paragraph{Node-wise learnability posterior and path sampling.}
The tree structure alone does not specify \emph{where} to explore next. Inspired by multi-armed bandits~\cite{slivkins2019introduction,gao2025bohdi}, we maintain a Beta posterior for each node $ \mathcal{D}_{i,j}$, which will be initialized to $(1,1)$.
\begin{equation}
\mathcal{M}(\mathcal{D}_{i,j}) =\mathrm{Beta}(\alpha_{i,j},\beta_{i,j})
\end{equation}
where $\mathcal{M}(\mathcal{D}_{i,j})$ represents the node-wise learnability, i.e. the probability that sampling the subtree rooted at $\mathcal{D}_{i,j}$ yields a \emph{learnable} training instance. We use Thompson sampling for path selection. At each depth, we sample a score from each candidate child's Beta posterior and choose the branch with the highest sampled score, which naturally trades off exploiting high-learnability regions and exploring uncertain ones.

\paragraph{Unk-triggered expansion.}
Since the fine-grained decomposition of a domain cannot be exhaustively enumerated in advance, we trigger incremental growth when sampling selects the unk node $\mathcal{D}_{i+1,\mathrm{unk}}$ under a parent node $\mathcal{D}_{i,j}$. Specifically, we generate a new sibling subtopic by
\begin{equation}
\mathcal{D}_{i+1,\mathrm{new}}=\mathrm{Expand}\!\big(\mathcal{D}_{i,j},\,\Psi(\mathcal{D}_{i,j})\big),
\end{equation}

where $\Psi(\mathcal{D}_{i,j})=\{\mathcal{D}_{i+1,1},\dots,\mathcal{D}_{i+1,N_{i+1}}\}$ denotes the set of existing children of $\mathcal{D}_{i,j}$ and serves as same-level context to encourage complementary, non-duplicate labels. We enforce deduplication by filtering $\mathcal{D}_{i+1,\mathrm{new}}\notin \Psi(\mathcal{D}_{i,j})$.

\paragraph{Web-backed validation for non-leaf nodes.}

To suppress hallucinated concepts and semantic drift, we validate non-leaf expansions ($i+1<\mathcal{L}$) using an external function $V_{\mathrm{wiki}}(\cdot)$. Given $\mathcal{D}_{i+1,\mathrm{new}}$, the validator returns a set of retrieved Wikipedia titles $\hat{W}$ and computes the maximum string-match similarity:
\begin{equation}
s_{\max}=\max_{w\in \hat{W}} \ \mathrm{sim}_{\mathrm{str}}\!\big(\mathcal{D}_{i+1,\mathrm{new}},\, w\big).
\end{equation}
If $s_{\max}\ge \tau_{\mathrm{wiki}}$, we add $\mathcal{D}_{i+1,\mathrm{new}}$ to the tree and create its next-layer unknown placeholder $\mathcal{D}_{i+2,\mathrm{unk}}$ to enable further growth; otherwise, we reject the expansion and re-sample to prevent early erroneous concept pollution of the tree.

\subsection{Open-Web Retrieval, Filtering, and Corpus Construction}
While the tree $\mathcal{T}$ provides a structured concept space, turning it into a trainable environment requires continuously attaching path-consistent external corpus to leaf concepts. Therefore, we maintain an external corpus pool for each leaf node $\mathcal{D}_{\mathcal{L},k}$.
\begin{equation}
\resizebox{0.85\linewidth}{!}{$
\begin{aligned}
\mathcal{C}(\mathcal{D}_{\mathcal{L},k})=\Big\{ F_{\mathrm{clean}}(p)\ \Big|\ & p\in\mathcal{U}_{\mathcal{L}, k},\ F_{\mathrm{url}}(p)=1,\\
& \cos\!\big(\phi(\mathcal{D}_{\mathcal{L},k}),\phi(H_p)\big)\ge \tau_{\mathrm{emb}}
\Big\}
\end{aligned}
$}
\end{equation}
where $\mathcal{U}_{\mathcal{L},k}$ is the set of URLs retrieved by querying the web with the leaf node $\mathcal{D}_{\mathcal{L},k}$ as the search keyword, $H_p$ is the page title, $\phi(\cdot)$ is a semantic encoder, $F_{\mathrm{url}}$ applies URL allow/deny lists to reduce noise and mitigate benchmark leakage, and $F_{\mathrm{clean}}$ removes boilerplate such as navigation bars, ads, templates, and duplicate blocks. This procedure converts the open-web into a leaf-aligned, continuously refreshed corpus environment for self-play training.

\subsection{Web-grounded Two Roles Self-Play Training}\label{sec:selfplay_cn}
At each iteration, we sample $B$ paths $\{\mathcal{P}_i\}_{i=1}^{B}$ from the tree $\mathcal{T}$. For each path $\mathcal{P}_i$, we sample a document $d$ from the corresponding leaf corpus pool $\mathcal{C}(\mathcal{D}_{\mathcal{L},i})$, and use a single policy $\pi_\theta$ to play both roles: \textbf{Challenger} generates QA pairs conditioned on the visible document $d$, and \textbf{Solver} answers the questions generated by the Challenger without access to $d$. Training signals come from verifiable rewards (RLVR), and policy updates are performed with Dr.\ GRPO.

\paragraph{Challenger: QA generation with verifiability filtering.}
For each document $d$, Challenger proposes $G$ candidate QA pairs $\{(q_i, y_i^*)\}_{i=1}^{G}\sim \pi_\theta(\cdot\mid d)$. We then apply a rule-based validator $\Gamma(\cdot)$ to filter out unverifiable or malformed instances, yielding
\begin{equation}
\mathcal{Q}(d)=\{(q_i,y_i^*)\mid \Gamma(q_i,y_i^*)=1\}.
\end{equation}
Each invalid QA will receive a penalty reward, discouraging malformed or unverifiable generations.

\paragraph{Solver: answer generation and solvability estimation.}
For a valid QA $(q_i,y_i^*)\in\mathcal{Q}(d)$, Solver answers the question $q_i$ by sampling $K$ independent responses: $\{\hat{y}_{i,k}\}_{k=1}^{K} \sim \pi_\theta(\cdot\mid q_i)$ and uses Math-Verify $v(\cdot,\cdot)$ to obtain correctness indicators $\ell_{i,k}=v(\hat{y}_{i,k},y_i^*)\in\{0,1\}$. We summarize solvability by the empirical accuracy and variance:
\begin{equation}
\hat{p}_i=\frac{1}{K}\sum_{k=1}^{K}\ell_{i,k},\ \widehat{\mathrm{Var}}_i=\hat{p}_i(1-\hat{p}_i).
\end{equation}
If all valid QA pairs satisfy $\widehat{\mathrm{Var}}_i=0$, then the document-level training signal is typically too easy, too hard, or unreliable. We skip policy updates for this document.

\paragraph{Reward design and Dr.\ GRPO updates.}
For a valid QA $(q_i,y_i^*)$, Solver will receive verifiable correctness rewards for each response:
\begin{equation}
r^{\mathrm{sol}}_{i,k}=\ell_{i,k}=v(\hat{y}_{i,k},y_i^*), k=1,\dots,K
\end{equation}
We assign a difficulty-shaped reward based on Solver's variance $\widehat{\mathrm{Var}}_i$, which peaks at moderate difficulty ($\hat{p}_i = 0.5$, i.e., $\widehat{\mathrm{Var}}_i = 0.25$) and decreases toward the extremes:
\begin{equation}
r^{\mathrm{cha}}_i=
\begin{cases}
\exp\!\left(-\frac{(\widehat{\mathrm{Var}}_i-0.25)^2}{\sigma}\right), & (q_i,y_i^*)\in\mathcal{Q}(d),\\
\rho, & \text{otherwise},
\end{cases}
\end{equation}
where $\sigma$ controls the width of the medium-difficulty band and $\rho<0$ penalizes invalid QA. 

\paragraph{Role balancing.} To keep the amount of training data aligned across the two roles, we uniformly sample one QA from the valid set, $(q^*,y^*)\sim \mathrm{Unif}(\mathcal{Q}(d))$, and use only this QA to construct Solver's grouped trajectories (i.e., $K$ Solver responses and their rewards). Challenger, in contrast, uses all $G$ QA pairs (valid with shaped rewards and invalid with punishment $\rho$) as its grouped samples.

\subsection{Posterior-guided Tree Updating with Sliding Window}
We convert self-play outcomes into feedback on the domain tree, closing the exploration--learning loop. For each valid QA of the sampled path $\mathcal{P}$, we define a Bernoulli learnability observation
\begin{equation}
g \triangleq \mathbf{1}\!\left(\widehat{\mathrm{Var}} \ge \tau_{\mathrm{var}}\right),
\end{equation}
where $\tau_{\mathrm{var}}\in(0,0.25]$ controls the width of the band around the capability boundary. Intuitively, $g=1$ indicates that the QA is likely near the current boundary and thus training-effective, whereas $g=0$ suggests that it is too easy, too hard, or unreliable. We attribute this feedback to all nodes on the $\mathcal{P}$ and perform Beta--Bernoulli conjugate updates:
\begin{equation}
\left\{
\begin{aligned}
\alpha_{i,j} &\leftarrow \alpha_{i,j}+g,\\
\beta_{i,j}  &\leftarrow \beta_{i,j}+(1-g),
\end{aligned}
\right.
\forall\, \mathcal{D}_{i,j}\in \mathcal{P}.
\end{equation}
Because learnability is non-stationary as the policy improves, accumulating statistics over the full history can bias exploration toward early observations. To mitigate this effect, we use a sliding window of the most recent $\mu$ observations per node to form effective parameters for sampling:
\begin{equation}
\left\{
\begin{aligned}
\tilde{\alpha}_{i,j} &= 1+\sum_{\tau\in \mathcal{W}_{i,j}} g^{(\tau)},\\
\tilde{\beta}_{i,j}  &= 1+\sum_{\tau\in \mathcal{W}_{i,j}} \bigl(1-g^{(\tau)}\bigr).
\end{aligned}
\right.
\end{equation}
where $\mu$ is window size, $\mathcal{W}_{i,j}$ denotes the indices of the most recent $\mu$ updates of $D_{i,j}$. During path sampling, we run Thompson sampling with $\mathrm{Beta}(\tilde{\alpha}_{i,j},\tilde{\beta}_{i,j})$, so that exploration preferences reflect the learnability distribution at the \emph{current} capability stage, improving both adaptivity and exploration efficiency.

\begin{table*}[t]
\centering
\footnotesize              
\setlength{\tabcolsep}{1.2pt}  
\renewcommand{\arraystretch}{1.1} 
\caption{\textbf{Main results} on mathematical and general reasoning benchmarks across four backbones. Best and second-best results within each backbone block are marked in \textbf{bold} and \underline{underline}, respectively.}
\label{tab:main}
\resizebox{0.9\textwidth}{!}{%
    \begin{tabular}{l|ccccccc|cccc|c}
    \toprule
    & \multicolumn{7}{c|}{\textbf{Mathematical Reasoning}} & \multicolumn{4}{c|}{\textbf{General Reasoning}} & \\
    \cmidrule(lr){2-8}\cmidrule(lr){9-12}
    \textbf{Method} &
    \textbf{AMC} & \textbf{Minerva} & \makecell{\textbf{MATH}\\\textbf{500}} & \textbf{GSM8K} & \textbf{Olymp.} &
    \makecell{\textbf{AIME}\\\textbf{24}} & \makecell{\textbf{AIME}\\\textbf{25}} &
    \makecell{\textbf{Super-}\\\textbf{GPQA}} & \makecell{\textbf{GPQA-}\\\textbf{Diamond}} & \makecell{\textbf{MMLU-}\\\textbf{Pro}} & \textbf{BBEH} &
    \textbf{Overall} \\
    \midrule
    
    \rowcolor{gray!10}\multicolumn{13}{c}{\textit{Qwen3-4B-Base}}\\
    Base Model                     & 41.4 & 35.7 & 57.0 & 75.9 & 30.2 & 9.5 &  6.4 & 18.0 & 32.8 & 51.5 &  8.2 & 33.3 \\
    \hspace{1em}+ R-Zero            & \underline{53.5} & 44.1 & \underline{77.0} & 91.1 & 39.5 & 10.3 &  7.1 & 26.7 & 33.4 & 53.7 & 10.4 & 40.6 \\
    \hspace{1em}+ SPICE     & 50.1 & \textbf{47.8} & 76.2 & \underline{92.5} & \textbf{41.0} & \textbf{12.0} & \textbf{10.9} & \underline{27.8} & \underline{35.1} & \underline{54.3} &  \textbf{11.8} & \underline{41.8} \\
    \rowcolor{blue!10}
    \textbf{\hspace{1em}+ WIST (ours)} & \textbf{60.0} & \textbf{47.8} & \textbf{78.2} & \textbf{92.9} & \underline{40.0} & \underline{11.6} & \underline{9.7} &
    \textbf{29.6} & \textbf{37.2} & \textbf{55.7} & \textbf{11.8} & \textbf{43.1} \\
    \midrule
    
    \rowcolor{gray!10}\multicolumn{13}{c}{\textit{Qwen3-8B-Base}}\\
    Base Model                     & 57.2 & 43.0 & 73.0 & 91.3 & 40.5 & 11.7 & 11.3 & 28.3 & 34.8 & 58.2 & 9.1 & 42.1 \\
    \hspace{1em}+ R-Zero            & \underline{61.1} & 48.5 & 80.4 & 92.9 & \underline{45.2} & 14.0 & 12.8 & \underline{31.8} & \textbf{42.4} & 60.4 & 11.2 & 45.5 \\
    \hspace{1em}+ SPICE     &60.1&\underline{51.5}&\underline{81.8}&\textbf{93.9}&\textbf{45.3}&\textbf{15.4}&\underline{13.4}&31.3&40.9&\underline{60.8}&\underline{11.6}&\underline{46.0} \\
    \rowcolor{blue!10}
    \textbf{\hspace{1em}+ WIST (ours)} & \textbf{63.4}&\textbf{53.3}&\textbf{82.6}&\underline{93.4}&44.1&\underline{14.8}&\textbf{13.9}&\textbf{32.5}&\underline{41.4}&\textbf{61.1}&\textbf{12.9}&\textbf{46.7} \\
    \midrule\midrule
    
    \rowcolor{gray!10}\multicolumn{13}{c}{\textit{OctoThinker-3B-Hybrid-Base}}\\
    Base Model                    & 12.5&18.5&30.6&44.9&11.0&1.7&\underline{0.6}&10.4&2.0&11.1&2.3&13.2 \\
    \hspace{1em}+ R-Zero            & 26.2&21.8&\underline{50.4}&73.5&\underline{17.2}&1.8&0.4&12.6&20.9&18.7&\underline{4.4}&22.5 \\
    \hspace{1em}+ SPICE     & \textbf{28.3}&\underline{22.4}&\textbf{50.8}&\textbf{76.7}&\textbf{17.3}&\textbf{2.7}&\textbf{0.8}&\textbf{18.4}&\underline{23.7}&\textbf{31.7}&\textbf{4.8}&\textbf{25.2} \\
    \rowcolor{blue!10}
    \textbf{\hspace{1em}+ WIST (ours)} & \underline{27.4}&\textbf{22.7}&48.8&\underline{76.3}&15.1&\underline{1.9}&\underline{0.6}&\underline{17.8}&\textbf{24.1}&\underline{30.4}&4.1&\underline{24.5} \\
    \midrule
    
    \rowcolor{gray!10}\multicolumn{13}{c}{\textit{OctoThinker-8B-Hybrid-Base}}\\
    Base Model                     & 20.0&26.2&42.8&82.2&17.0&2.4&1.1&16.4&12.1&25.9&5.4&22.9 \\
    \hspace{1em}+ R-Zero            & 25.2&\underline{31.5}&\underline{58.7}&86.3&\textbf{25.9}&\underline{3.5}&\textbf{1.5}&\underline{24.1}&\underline{27.3}&\underline{42.5}&9.9&30.6 \\
    \hspace{1em}+ SPICE     & \textbf{33.8}&30.2&58.6&\textbf{87.6}&24.9&\textbf{4.8}&0.9&23.3&\textbf{30.8}&40.5&\textbf{10.4}&\underline{31.4} \\
    \rowcolor{blue!10}
    \textbf{\hspace{1em}+ WIST (ours)} & \underline{31.0}&\textbf{36.4}&\textbf{62.0}&\underline{87.0}&\underline{25.5}&3.2&\underline{1.4}&\textbf{25.4}&30.6&\textbf{45.9}&\underline{10.1}&\textbf{32.6} \\
    \bottomrule
    \end{tabular}
}
\end{table*}

\section{Experiments}

\subsection{Setup}

\paragraph{Models and baseline.} Following R-Zero and SPICE, we evaluate WIST on two model families and scales: \textit{Qwen3-4B-Base}/\allowbreak\textit{Qwen3-8B-Base}~\cite{yang2025qwen3} and \textit{OctoThinker-3B}/\allowbreak\textit{OctoThinker-8B}~\cite{wang2025octothinker}. 
We compare WIST against the following baselines: (1) \textbf{Base Model}: the pretrained model without any post-training, serving as the performance floor; (2) \textbf{R-Zero}: a fully endogenous self-play method that relies only on prompting and self-generated problems, without accessing external data; (3) \textbf{SPICE}: a corpus-grounded self-play method that uses a curated high-quality pretraining corpus (Nemotron-CC-Math~\cite{mahabadi2025nemotron}) as the environment. All baseline implementations are provided in the Appendix~\ref{sec:training_details}.

\paragraph{Evaluation Benchmarks.}
We evaluate WIST on a broad suite of math and general reasoning benchmarks, largely following the setups in R-Zero and SPICE, and additionally include a physics benchmark to test domain-specific gains.
(1) \textbf{Mathematical reasoning.} We report results on AMC, Minerva~\cite{lewkowycz2022solving}, MATH-500~\cite{hendrycks2021measuring}, GSM8K~\cite{cobbe2021training}, OlympiadBench~\cite{he2024olympiadbench}, AIME'24, and AIME'25.
We report accuracy based on greedy decoding for most evaluations, following~\cite{ma2025general}. The only exceptions are AIME'24 and AIME'25, where scores are averaged across 32 sampling runs as in~\cite{zeng2025simplerl}. 
(2) \textbf{General-domain reasoning.} To measure generalization beyond math, we evaluate on MMLU-Pro~\cite{wang2024mmlu}, SuperGPQA~\cite{du2025supergpqa}, GPQA-Diamond~\cite{rein2024gpqa}, and BBEH~\cite{kazemi2025big}, following the prompts and evaluation code from~\cite{ma2025general} and reporting accuracy under greedy decoding. Detailed evaluation settings are provided in Appendix~\ref{sec:evaluation_settings}.

\paragraph{Training Details.}

Our entire framework is implemented based on the OpenRLHF codebase~\cite{hu2024openrlhf} and set the target domain to \textbf{Mathematics}, consistent with prior self-evolution studies such as R-Zero and SPICE. In each iteration, we sample $B=128$ root-to-leaf paths from the domain tree with maximum depth $\mathcal{L}=4$.
We define the learnability event using self-consistency variance and use a threshold $\tau_{\mathrm{var}}=0.2$, which is consistent with the range used in prior work~\cite{zhang2025consistent,huang2025r,bercovich2025llama}, with a sliding window of size $\mu=5$ for posterior updates.
For open-web corpus acquisition, we filter retrieved pages by title semantic similarity with threshold $\tau_{\mathrm{emb}}=0.5$, and truncate each cleaned document to at most 5992 tokens.
For self-play, Challenger and Solver both repeat sampling G=8 times; invalid QA candidates receive a fixed penalty $\rho=-0.1$.
We optimize the policy with Dr.GRPO using training batch size $B_{\mathrm{train}}=512$. 
All other hyperparameters and implementation details are provided in Appendix~\ref{sec:training_details}.

\begin{figure*}[t]
	\centering
	\begin{subfigure}[t]{0.32\textwidth}
		\centering
		\includegraphics[width=\linewidth]{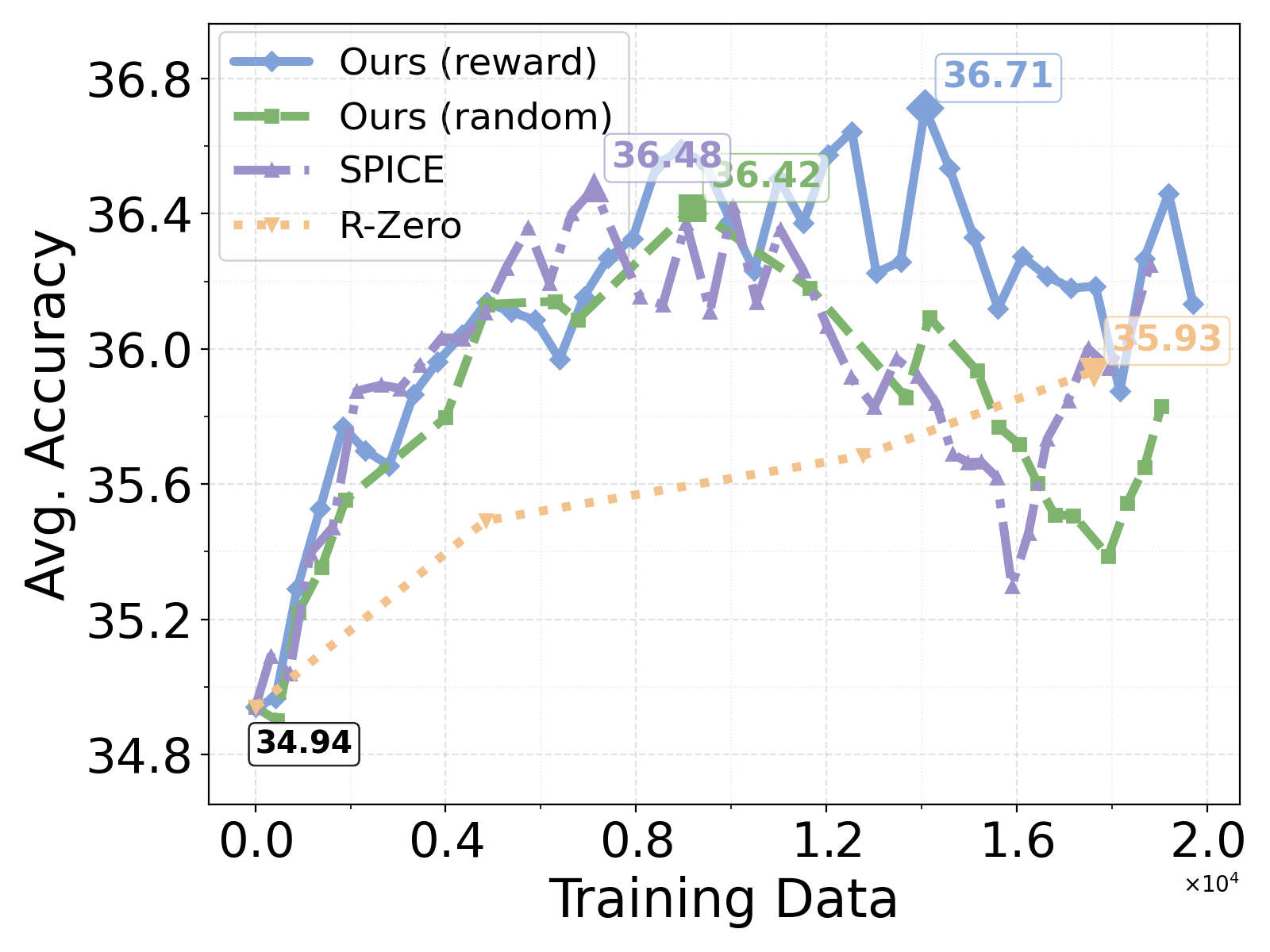}
		\caption{Qwen3-4B-Base}
		\label{fig:qwen3-4b-base-med}
	\end{subfigure}\hfill
	\begin{subfigure}[t]{0.32\textwidth}
		\centering
		\includegraphics[width=\linewidth]{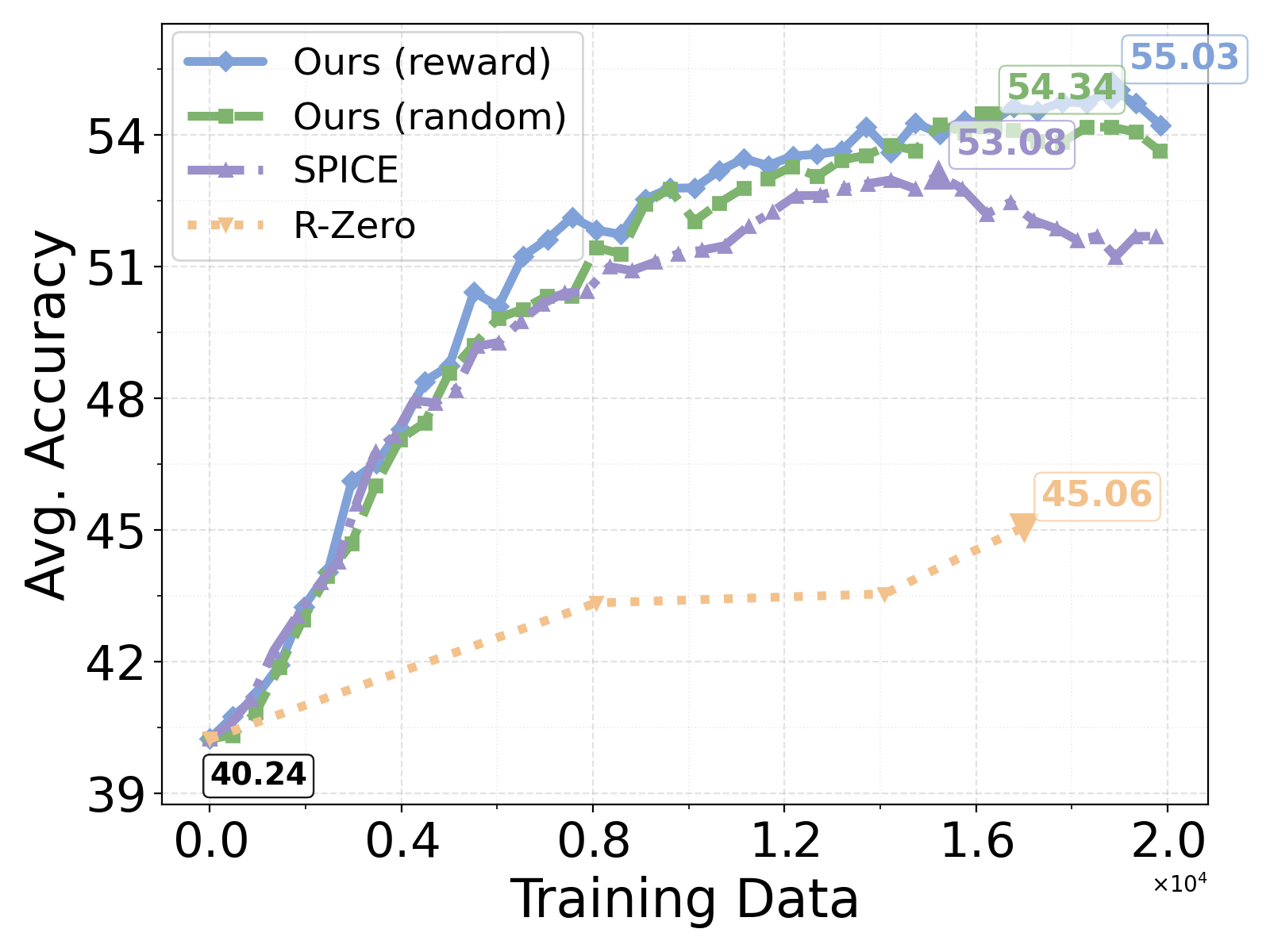}
		\caption{Qwen3-8B-Base}
		\label{fig:qwen3-8b-base-med}
	\end{subfigure}\hfill
	\begin{subfigure}[t]{0.32\textwidth}
		\centering
		\includegraphics[width=\linewidth]{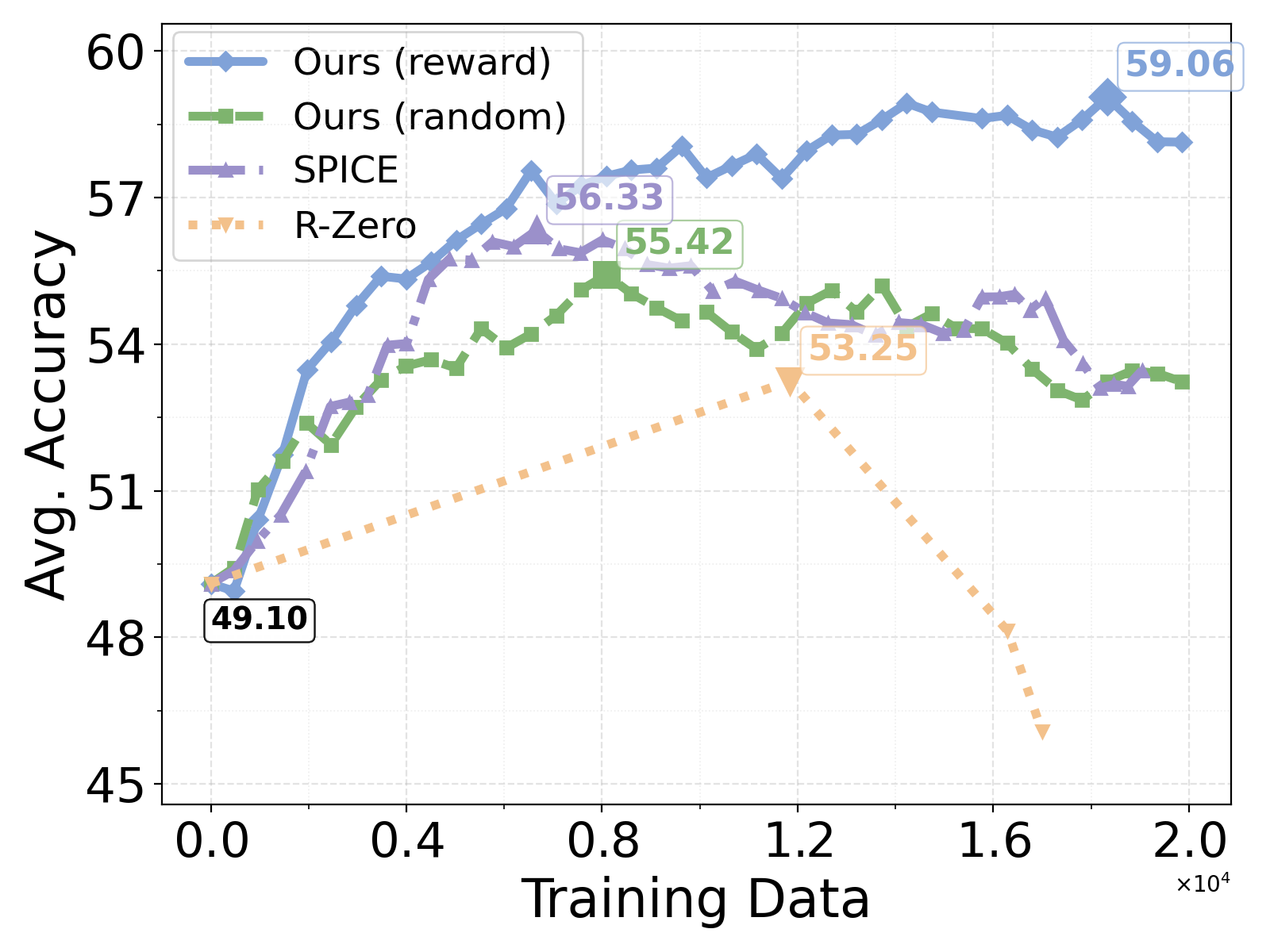}
		\caption{Qwen3-14B-Base}
		\label{fig:qwen3-14b-base-med}
	\end{subfigure}
	
	\caption{Training performance of our method \textbf{WIST} in the medical domain using three Qwen3 models of different sizes.}
	\label{fig:wist-qwen3-med}
\end{figure*}

\subsection{Main Results}
As shown in Table~\ref{tab:main}, WIST consistently outperforms the base models across all four backbones and achieves the best overall performance in three of them. Specifically, on \textit{Qwen3-4B-Base}, WIST improves the Overall score from 33.3 to 43.1 (+9.8), surpassing R-Zero (40.6, +7.3) and SPICE (41.8, +8.5). This indicates that, even without relying on a carefully curated corpus, WIST can continuously produce effective training signals through structured exploration and web-grounded corpus construction. On \textit{Qwen3-8B-Base}, WIST again attains the highest Overall score (46.7), achieving a +4.6 gain over the base model and outperforming both R-Zero (45.5) and SPICE (46.0). This suggests that as model capacity increases, WIST's exploration--retrieval--self-play loop translates more reliably into cross-task generalization gains. Moreover, on \textit{OctoThinker-8B-Hybrid-Base}, WIST raises the Overall score to 32.6 (+9.7), exceeding R-Zero (30.6) and SPICE (31.4). These results corroborate that when the underlying model has sufficient reasoning and information-integration capability, WIST's tree-structured decomposition and posterior-guided exploration can more effectively identify weaknesses, broaden long-tail concept coverage, and yield cumulative improvements on both mathematical and general reasoning benchmarks. Additional results on a larger-scale model (\textit{Qwen3-14B-Base}) are provided in Appendix~\ref{sec:qwen14b_appendix}.

In contrast, under \textit{OctoThinker-3B-Hybrid-Base}, WIST achieves an Overall score of 24.5, slightly below SPICE's 25.2, while still substantially outperforming R-Zero (22.5) and the base model (13.2). This outcome is expected: WIST relies on open-web retrieval and automatic cleaning to construct its corpus environment. Although relevance filtering and controllability constraints reduce noise, the open web inevitably contains noisy, ambiguous, or weakly related content. For a smaller 3B model, such noise can more easily amplify misleading gradients and distributional drift during self-play, thereby undermining training stability. By contrast, SPICE operates in a carefully curated in-corpus environment with more controlled data quality and distribution, which mitigates the adverse impact of noise in the small-model regime. Overall, the results indicate that WIST better realizes the benefits of open-web corpus and tree-guided curricula for medium-to-large models, while in the small-model setting, the interaction between environmental noise and limited model capacity becomes a key factor shaping the attainable gains.

\begin{figure*}[t]
	\centering
	
	\begin{subfigure}[t]{0.245\textwidth}
		\centering
		\includegraphics[width=\linewidth]{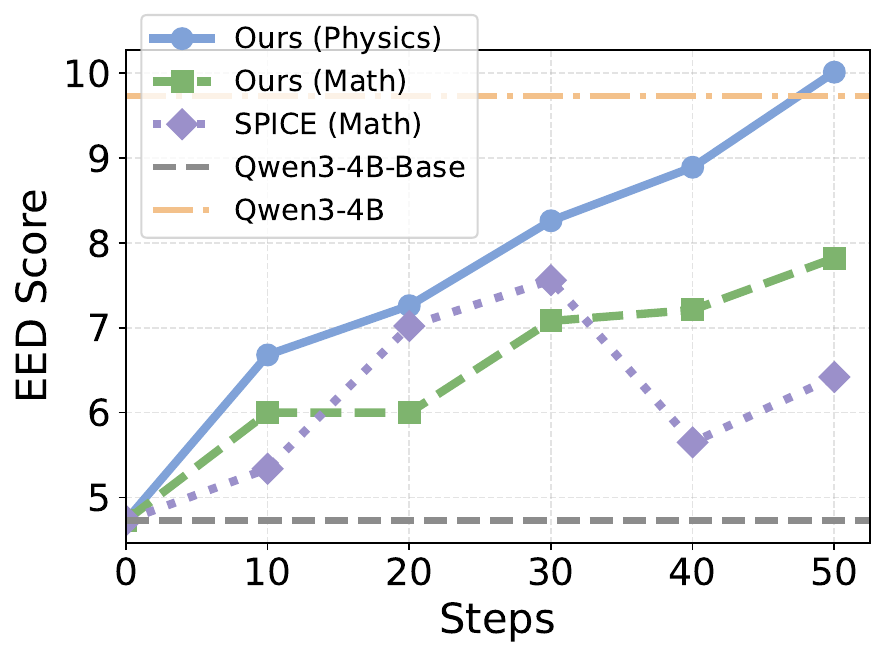}
		\caption{PhyBench EED scores over training steps.}
		\label{fig:phybench_eed}
	\end{subfigure}\hfill
	\begin{subfigure}[t]{0.22\textwidth}
		\centering
		\includegraphics[width=\linewidth]{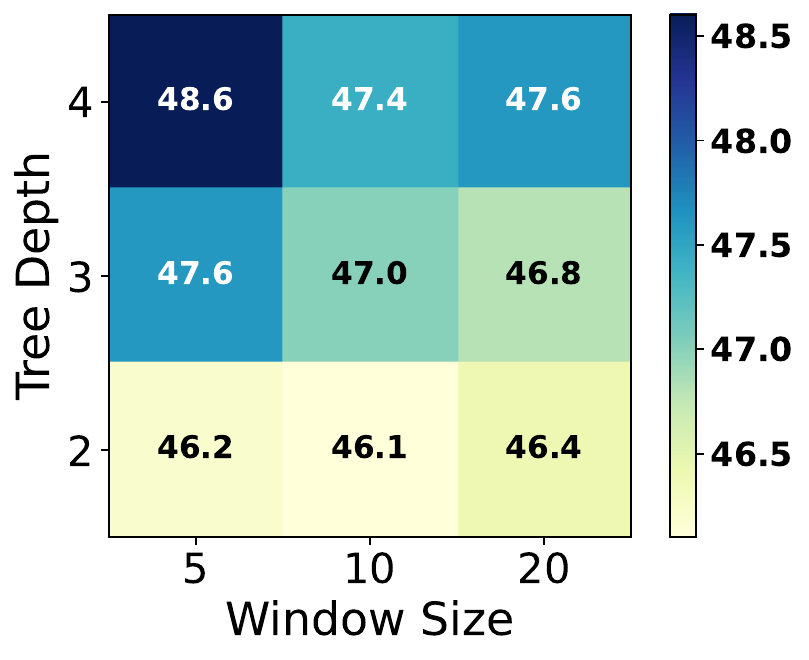}
		\caption{Math average performance across tree-depth and window-size settings.}
		\label{fig:levels_window_heatmap}
	\end{subfigure}\hfill
	\begin{subfigure}[t]{0.24\textwidth}
		\centering
		\includegraphics[width=\linewidth]{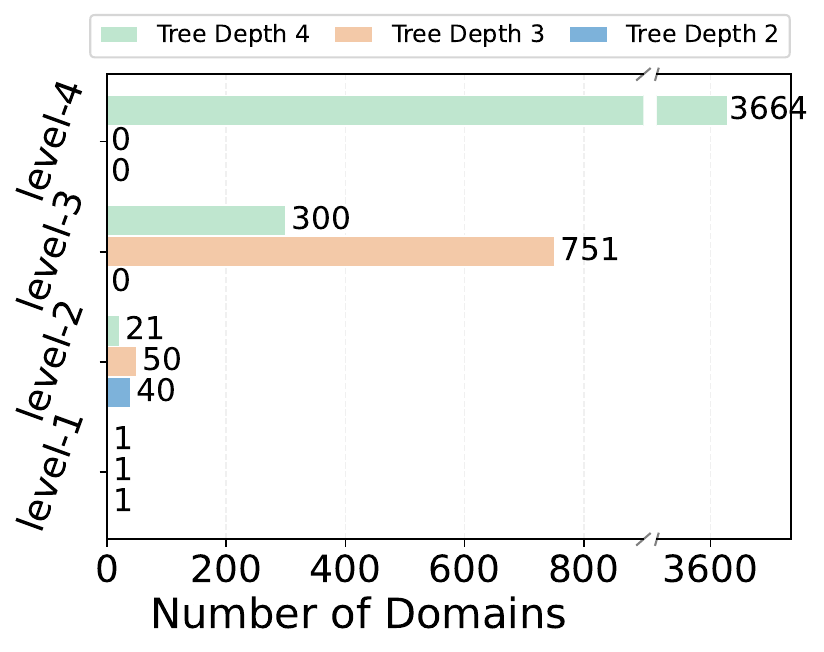}
		\caption{Number of nodes under different tree depths.}
		\label{fig:tree_nodes}
	\end{subfigure}\hfill
	\begin{subfigure}[t]{0.25\textwidth}
		\centering
		\includegraphics[width=\linewidth]{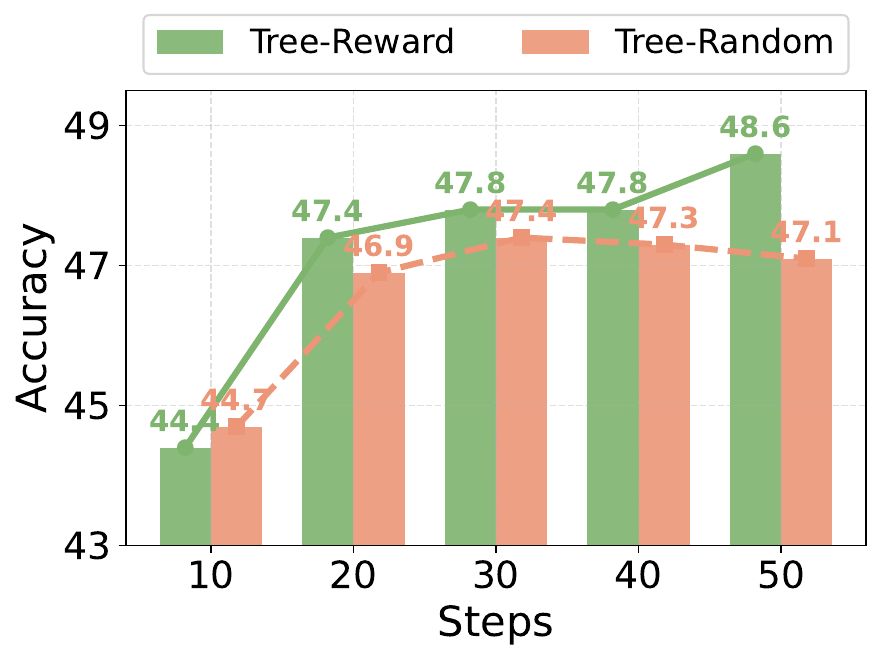}
		\caption{Random vs. Reward-Guided Exploration: Math average performance over Steps.}
		\label{fig:random_vs_reward}
	\end{subfigure}
	
	\caption{\textbf{Ablation results of WIST}, including domain transfer, hyperparameter sensitivity, tree scaling, and exploration strategy.}
	\label{fig:wist_analysis_four}
\end{figure*}

\section{Ablation Studies}
\subsection{Domain Transfer Beyond Mathematics}
\label{abl_domian}

To test whether WIST can be steered beyond mathematics, we study transfer to two scientific domains, \textbf{medicine} and \textbf{physics}. We report broader experiments in medicine on three Qwen3 base models, and a focused study on \textit{Qwen3-4B-Base} in physics. This difference mainly reflects the SPICE setup: high-quality curated corpora are relatively abundant in medicine, but much more limited in physics.

\paragraph{Transfer to medicine.}
We steer the target domain to \textbf{medicine} and evaluate WIST on \textit{Qwen3-4B-Base}, \textit{Qwen3-8B-Base}, and \textit{Qwen3-14B-Base}. For comparison, we use a PubMed-based medical corpus from prior work~\cite{kandpal2025common} as the domain corpus for SPICE. We use OpenCompass~\cite{2023opencompass} for evaluation and report the average accuracy on three medical benchmarks:

\begin{itemize}
    \item \textbf{Medbullets}~\cite{chen2025benchmarking}: a clinically oriented benchmark with USMLE-style questions and expert explanations, testing grounded medical reasoning and clinical decision-making.
    \item \textbf{MedMCQA}~\cite{pmlr-v174-pal22a}: a large-scale multiple-choice benchmark built from real-world medical entrance examination questions.
    \item \textbf{MedQA}~\cite{jin2020disease}: an exam-based multiple-choice benchmark collected from multiple regions and accompanied by medical textbooks, suitable for evaluating professional medical knowledge and reasoning.
\end{itemize}

As shown in Figure~\ref{fig:wist-qwen3-med}, WIST consistently improves medical-domain performance across all three model sizes and outperforms both SPICE and R-Zero. It raises average accuracy by \textbf{+1.77} on \textit{Qwen3-4B-Base}, \textbf{+14.79} on \textit{Qwen3-8B-Base}, and \textbf{+9.96} on \textit{Qwen3-14B-Base}. We also observe larger gains on larger models: compared with the modest improvement on 4B, both 8B and 14B benefit substantially more, suggesting that higher-capacity models can make better use of retrieved medical evidence and convert it into stronger reasoning gains. The detailed results of the best checkpoints for each model are provided in the appendix~\ref{sec:medical_results}.

\paragraph{Transfer to physics.}
We also steer the target domain to \textbf{physics} and evaluate WIST on \textit{Qwen3-4B-Base}. We still use OpenCompass for evaluation on \textbf{PhyBench}~\cite{qiu2025phybench}, a physics reasoning benchmark that requires models to generate structured LaTeX expressions. We report the \textbf{EED Score} (Expression Edit Distance Score; range 0--100), which compares predicted and reference expressions through expression-tree alignment and captures structural and semantic partial correctness. For stability, each evaluation is repeated 10 times and we report the mean score. In this setting, because a comparable high-quality curated \textbf{physics corpus} is not readily available, SPICE still uses its mathematics corpus.

As shown in Figure~\ref{fig:phybench_eed}, WIST yields clear gains on PhyBench. For example, \textit{Qwen3-4B-Base} improves from 4.73 to 10.01 after 50 training steps, outperforming both the base checkpoint and a stronger reasoning-enhanced baseline (\textit{Qwen3-4B} in thinking mode). This indicates that WIST can acquire domain-relevant corpus from the open web and translate it into measurable improvements in physics reasoning and symbolic expression generation.


\subsection{Coupled Sensitivity of Tree Depth and Window Size}

We next examine the interaction between two key hyperparameters, tree depth ($\mathcal{L}$) and window size ($\mu$). Intuitively, $\mathcal{L}$ controls the granularity and branching of the exploration space, while $\mu$ controls how quickly posterior guidance adapts to policy non-stationarity. Rather than tuning them independently, we perform a grid sweep over $\mathcal{L}\in\{2,3,4\}$ and $\mu\in\{5,10,20\}$, and evaluate the average performance of \textbf{Math} after training for 50 steps on the Qwen3-4B-Base model. Results are summarized in the heatmap in Figure~\ref{fig:levels_window_heatmap}.

We observe a clear coupling effect: shallower trees favor larger windows, whereas deeper trees favor smaller windows. This trend is consistent with how the effective search space scales with depth. As $\mathcal{L}$ increases, the number of reachable leaf concepts grows rapidly, increasing path diversity and reducing the revisit frequency of any single path. As shown in Figure~\ref{fig:tree_nodes}, the deeper the tree's depth, the more nodes explored, and the larger the search space. Under such high diversity, a large window may aggregate stale or heterogeneous feedback and introduce noise, while a smaller window better tracks local, recent learnability. Conversely, when $\mathcal{L}$ is small, paths are revisited more frequently, and larger windows provide more stable statistics for posterior estimation. 

Notably, setting the tree depth to 2 is effectively equivalent to directly prompting the model to generate minimal knowledge points within the target domain. As shown in Figure~\ref{fig:tree_nodes}, this configuration yields only about 40 leaf-level concepts at the second layer. Meanwhile, Figure~\ref{fig:levels_window_heatmap} indicates that without deeper, structured tree guidance, it is difficult for the model to reliably retrieve and organize high-value, domain-relevant knowledge from the vast and noisy open web. These results provide empirical corpus for the necessity and critical role of the proposed Tree component in open-web self-evolution.

%

\subsection{Benefit of Reward-Guided Tree Expansion}

We isolate the contribution of \textbf{reward-guided exploration} in tree construction. In this experiment, we train \textit{Qwen3-4B-Base} for 50 steps in the \textbf{mathematics} domain and compare two variants for selecting expansion paths and sampling branches: 
(1) \textbf{Random exploration}, where candidate nodes are selected uniformly at random without using any feedback signal; and 
(2) \textbf{Reward exploration}, where sampling is driven by the posterior updated from learnability feedback. 
As shown in Figure~\ref{fig:random_vs_reward}, which reports the average performance on the math benchmark over training steps, the two variants perform similarly at early stages but diverge as training proceeds. Reward-guided exploration yields a more stable improvement trajectory and achieves the best final performance at 50 steps, whereas random exploration remains consistently weaker. The detailed results at step 50 are provided in Appendix~\ref{sec:random_reward}. This suggests that unguided expansion tends to disperse effort across the enlarged search space and fails to consistently focus on high-yield regions. The same pattern is observed in the medical-domain experiments in Section~\ref{abl_domian}, where reward exploration consistently outperforms random exploration across different model sizes, as shown in Figure~\ref{fig:wist-qwen3-med}. Together, these results indicate that the effectiveness of WIST comes not merely from broader exploration, but from using learnability signals to direct exploration toward high-yield regions of the domain tree.



\section{Conclusion}

We proposed \textbf{WIST}, a web-grounded iterative self-play tree framework for domain-targeted reasoning improvement. WIST closes the loop between structured exploration and learning by expanding a domain tree, retrieving path-consistent web corpus, training a Challenger--Solver self-play process with verifiable rewards, and feeding learnability signals back to guide future exploration. Across diverse backbones, WIST consistently improves over the base models and typically outperforms both purely endogenous self-evolution (R-Zero) and corpus-grounded self-play (SPICE). WIST is also inherently domain-steerable: by switching only the target domain label, it achieves strong improvements in both \textbf{medicine} and \textbf{physics}, demonstrating effective transfer beyond mathematics. Ablations further confirm the effectiveness of WIST's key components for stable open-web learning. Overall, WIST combines the flexibility and coverage of open-web grounding with structured, feedback-guided exploration, providing a practical, scalable, and domain-adaptive route to self-improvement without relying on manually curated domain corpora.

\paragraph{Limitations.}

WIST builds on reinforcement learning with verifiable rewards (RLVR) and is most effective when correctness can be checked reliably at scale (e.g., mathematics). For expert-dependent domains with ambiguity and multiple acceptable answers, such as law and medicine, it is widely recognized that designing robust verifiers and obtaining consistently informative RLVR signals is difficult; we report exploratory attempts and discussion in the appendix. WIST also relies on open-web retrieval, where corpus quality and alignment can vary despite filtering and cleaning. This variability can attenuate gains for smaller models, but we still observe consistent improvements overall.


\bibliography{custom}

\appendix
\section{Algorithm Implementation}
\label{sec:algorithm}
We presented the overall framework of WIST in Algorithm~\ref{alg:wist}.

\newcommand{\rcomm}[1]{\hfill$\triangleright$~#1}
\begin{algorithm*}[t]
\caption{\textsc{WIST}: Web-grounded Iterative Self-play Tree}
\label{alg:wist}
\resizebox{0.9\textwidth}{!}{%
\begin{minipage}{\textwidth}
\begin{algorithmic}[1]
\Require Policy $\pi_{\theta_{0}}$; target domain $D_{1,1}$; depth $\mathcal{L}$; window $\mu$; iterations $T$; rollout batch $B$; group size $G$

\State Initialize domain tree $\mathcal{T}$ with virtual root $D_{0,0}$; add $D_{1,1}$ (and its “unk” child) to $\mathcal{T}$.
\State For each node $u$, keep Beta posterior (init $(1,1)$) with a sliding buffer of length $\mu$.

\For{$t \gets 1$ to $T$}
    \State \textbf{Part I: Path sampling and tree expansion.}
    \For{$b \gets 1$ to $B$}
        \State $P_b \gets [D_{0,0}]$; $u\gets D_{0,0}$
        \For{$i \gets 0$ to $\mathcal{L}-1$}
            \State Sample $s_c\sim\mathrm{Beta}(\tilde{\alpha}(c),\tilde{\beta}(c))$ for each child $c\in\mathrm{Ch}(u)$; set $c\gets\arg\max s_c$
            \While{$c=D_{i+1,\mathrm{unk}}$} \Comment{Unk-triggered expansion}
                \State $D_{i+1,\mathrm{new}}\gets \mathrm{Expand}(u,\Psi(u))$ \Comment{$\Psi(u)$: existing sibling labels}
                \If{$D_{i+1,\mathrm{new}}\in \Psi(u)$} \State \textbf{continue} \EndIf
                \If{$i+1<\mathcal{L}$} \Comment{non-leaf validation}
                    \State $s_{\max}\gets \max_{w\in V_{\mathrm{wiki}}(D_{i+1,\mathrm{new}})} \mathrm{sim}_{\mathrm{str}}(D_{i+1,\mathrm{new}},w)$
                    \If{$s_{\max}<\tau_{\mathrm{wiki}}$} \State \textbf{continue} \EndIf
                \EndIf
                \State Add $D_{i+1,\mathrm{new}}$ (and its ``unk'' child if $i+1<\mathcal{L}$) to $\mathcal{T}$; init Beta$(1,1)$
                \State $c\gets D_{i+1,\mathrm{new}}$
            \EndWhile
            \State Append $c$ to $P_b$; $u\gets c$
        \EndFor
        \State $\ell_b \gets u$ \Comment{leaf node of $P_b$}
        \State \textbf{Leaf corpus:} $\mathcal{C}(\ell_b)\gets \mathrm{RetrieveAndClean}(\ell_b)$
    \EndFor

    \State \textbf{Part II: Web-grounded two-roles self-play and updates.}
    \For{$b \gets 1$ to $B$}
        \State \textbf{Challenger:} propose $G$ QA pairs from $d\sim\mathcal{C}(\ell_b)$; invalid ones receive penalty $\rho$; valid set $\mathcal{Q}(d)$ is kept by $\Gamma(\cdot)$.
        \If{$\mathcal{Q}(d)=\emptyset$} \State \textbf{continue} \EndIf
        \State \textbf{Solver:} for each $(q,y^\star)\in\mathcal{Q}(d)$, sample $G$ answers and compute $\widehat{\mathrm{Var}}(q,y^\star)$ via $v(\cdot,\cdot)$.
        \If{$\forall(q,y^\star)\in\mathcal{Q}(d),\ \widehat{\mathrm{Var}}(q,y^\star)=0$} \State \textbf{continue} \EndIf
        \State \textbf{Rewards:} Solver uses correctness $r_S=v(\hat{y},y^\star)$; Challenger uses $r_C=\exp\!\left(-\frac{(\widehat{\mathrm{Var}}-0.25)^2}{\sigma}\right)$ for valid QA and $\rho$ otherwise.
        \State \textbf{Role balancing:} uniformly sample one valid QA $(q^\star,y^\star)\sim\mathrm{Unif}(\mathcal{Q}(d))$ to form the Solver training group.
        \State Use the Challenger group (size $G$) and the Solver group (size $G$) to separately calculate the advantage $\mathcal{A}$ of Dr. GRPO.
        \State \textbf{Posterior feedback:} $g\gets\mathbf{1}\!\left[\widehat{\mathrm{Var}}(q^\star,y^\star)\ge\tau_{\mathrm{var}}\right]$.
        \ForAll{$u\in P_b$} \State Append $g$ to buffer$(u)$ (keep last $\mu$) and recompute $(\tilde{\alpha}(u),\tilde{\beta}(u))$. \EndFor
    \EndFor
    Update the $\pi_{\theta_t-1}$ to $\pi_{\theta_{t}}$ using the advantage $\mathcal{A}$.
\EndFor
\State \Return trained $\pi_{\theta_{T}}$ and tree $\mathcal{T}$
\end{algorithmic}
\end{minipage}}
\end{algorithm*}

\begin{figure*}[t]
	\centering
	\begin{minipage}{0.9\textwidth}
		\small\ttfamily
		Root \hspace{2em} (virtual root at level 0)\\
		\hspace*{0.5em}`-- Mathematics \hspace{2em} (target domain at level 1)\\
		\hspace*{2em}|-- unk \hspace{6.7em} (placeholder for expansion at level 2)\\
		\hspace*{2em}`-- Calculus \hspace{4.8em} (sub-domain at level 2)\\
		\hspace*{4.5em}|-- unk \hspace{6.7em} (placeholder for expansion at level 3)\\
		\hspace*{4.5em}`-- Differential Equations \hspace{1.2em} (sub-domain at level 3)\\
		\hspace*{7.5em}|-- unk \hspace{10.2em} (placeholder for expansion at level 4)\\
		\hspace*{7.5em}|-- Method of Undetermined Coefficients \hspace{0.5em} (minimal knowledge point at level 4)\\
		\hspace*{7.5em}`-- Method of Variation of Parameters \hspace{1.7em} (minimal knowledge point at level 4)
	\end{minipage}
	\caption{\textbf{Example of a partial mathematics domain label tree generated by WIST.} Each non-leaf level contains an \texttt{unk} child that indicates a possible expansion point.}
	\label{fig:tree_example}
\end{figure*}

\section{Related Work}
\label{sec:related}
\paragraph{Reasoning-focused RL and verifiable supervision.}
Reinforcement learning has been widely used to align LLMs with human preferences~\cite{jaques2019way,ouyang2022training}, and more recently to directly enhance reasoning by optimizing rewards that can be checked automatically. RL with verifiable rewards (RLVR) has proven effective in domains with crisp correctness signals (e.g., math or executable programs), where rule-based or deterministic verification can replace expensive human judgments~\cite{uesato2022solving, lightman2023let}. A practical limitation is that many RLVR pipelines still depend on pre-collected tasks, which restrict coverage and domain adaptation. \emph{Our work complements RLVR by turning the open web into a continuously refreshable source of verifiable training instances, enabling domain-targeted improvement without relying on a pre-arranged domain corpus.}

\paragraph{Self-play and automatic curricula for language models.}
Self-play is a general mechanism for creating training curricula through interaction, and has long been central to game-playing systems~\cite{lightman2023let,Bakhtin2022HumanlevelPI}. In language modeling, self-play has been explored both for alignment (e.g., models critiquing or rewarding their own outputs) and for capability gains via dual-role or adversarial setups~\cite{chen2024self,yuan2024self,cheng2024self,liu2025spice,huang2025r}. However, applying multi-agent RL to full LLMs often requires simplifying assumptions or bespoke environments, and performance can hinge on how well the interaction setting controls task difficulty and data quality~\cite{harding-graesser-etal-2019-emergent,Sarkar2025TrainingLM}. \emph{Like SPICE~\cite{liu2025spice}, WIST falls under corpus-grounded self-play, but differs in both the environment and the training process. Rather than assuming a carefully curated domain corpus, WIST treats the open web as the self-play environment and organizes it with a dynamically expanding domain tree. Starting from only a user-specified target domain, WIST autonomously expands subtopics, retrieves path-consistent web evidence, and uses posterior-guided sampling to induce a learnability-aware curriculum during self-play.}

\paragraph{Endogenous self-evolution and label-free reward signals.}
A growing body of work seeks to reduce reliance on labeled data by deriving rewards from the model itself, such as confidence, entropy, or agreement across multiple samples~\cite{ouyang2022training,chen2025self}. These signals are often coupled with self-training loops that iteratively fine-tune on the model's own solutions~\cite{zhao2025learning, shafayat2025can}. Fully endogenous variants can even generate problems from scratch, but they may suffer from drift as errors compound and the generated distribution departs from grounded knowledge. \emph{Our approach reduces this failure mode by grounding the self-play loop in retrieved corpus and by constraining exploration through a structured decomposition of the target domain, which helps maintain coverage and training stability over iterations.}

\paragraph{Web/corpus mining and synthetic QA generation.}
Another line of research scales reasoning data by mining questions from documents or generating synthetic QA from prompts, either bootstrapping from existing datasets or harvesting from large corpora and the web~\cite{wang2023self,zelikman2022star,yu2023metamath,mahabadi2025nemotron,yuan2025naturalreasoning}. Most of these pipelines are offline: they produce static datasets whose distribution is fixed once collected, and they typically require extensive filtering rules to ensure quality. More interactive approaches generate questions online from document contexts to better match the learner's current capability, but often assume a curated corpus as the environment. \emph{WIST is distinct in that it does not require a fixed in-corpus environment: it continuously acquires corpus from the open web and couples retrieval with posterior-guided exploration over a dynamically expanded domain tree, enabling domain-steerable self-evolution with minimal manual data curation.}

\section{Example of a Domain Label Tree}
\label{sec:tree_example}

To improve readability, we provide here a concrete example of a partial domain label tree generated by WIST. Starting from a user-provided target domain label, WIST incrementally expands the tree into finer-grained subtopics, while reserving an \texttt{unk} child at each non-leaf level to indicate a possible expansion point. 

Figure~\ref{fig:tree_example} shows a small real snapshot of a domain tree rooted at \textbf{Mathematics}. First, path sampling selects a root-to-leaf route, for example, \textit{Mathematics} $\rightarrow$ \textit{Calculus} $\rightarrow$ \textit{Differential Equations} $\rightarrow$ \textit{Method of Undetermined Coefficients}. Second, if sampling selects an \texttt{unk} node, WIST expands the tree by generating a new sibling label under the corresponding parent node. Third, after self-play on documents associated with the selected leaf, the resulting learnability signal is used to update the node statistics along the sampled path, which in turn influences future path sampling.

\section{Training Details}
\label{sec:training_details}
For R-Zero, we run the official released implementation with the same backbone models and compute budget whenever applicable. Since SPICE has not publicly released its code, we have reproduced the baseline based on the algorithmic process described in its paper. To reduce confounding factors, we align the optimizer, sampling strategy, and training steps with those used by WIST whenever possible, and we report ablations to quantify the effect of each additional component beyond the shared training recipe.

\paragraph{Baseline hyperparameters.} All the overall hyperparameters of the comparison methods are shown in the Table~\ref{setting_baselines}.

\begin{table*}[t]
\centering
\small
\setlength{\tabcolsep}{7pt}
\renewcommand{\arraystretch}{1.15}
\caption{Training configurations for all compared methods.}
\label{setting_baselines}
\begin{tabularx}{\textwidth}{l *{4}{>{\centering\arraybackslash}X}}
\toprule
\textbf{Configuration} & \textbf{WIST} & \textbf{SPICE} & \textbf{R-Zero}  \\
\midrule

\multicolumn{4}{l}{\textit{Data Source}} \\
Corpus documents    & $\sim15,000$ & 20,000 & \textendash  \\
Question source     & Web-grounded & Document-grounded & Self-generated  \\
External grounding  & $\checkmark$ & $\checkmark$ & $\times$  \\
\midrule

\multicolumn{4}{l}{\textit{Training Details}} \\
Challenger training & $\checkmark$ & $\checkmark$ & $\checkmark$  \\
Challenger sampling & 8 & 8 & 4  \\
Reasoner training   & $\checkmark$ & $\checkmark$ & $\checkmark$ \\
Reasoner sampling   & 8 & 8 & 5  \\
Temperature         & 1.0 & 1.0 & 1.0 \\
Optimizer           & DrGRPO & DrGRPO & GRPO \\
\midrule

\multicolumn{4}{l}{\textit{Reward Design}} \\
Challenger reward   & Gaussian Variance (max 1.0) & Gaussian Variance (max 1.0) & $1-2|p-0.5|$  \\
Reasoner reward     & Binary correctness & Binary correctness & Binary correctness \\
Invalid penalty     & -0.1 & -0.1 & -1 (Challenger) \\
\midrule

\multicolumn{4}{l}{\textit{Performance}} \\
Training iterations & 50 & 50 & $3^{\dagger}$ \\
\bottomrule
\end{tabularx}
\end{table*}

\paragraph{WIST hyperparameters.}
Table~\ref{tab:wist_hparams} lists the main hyperparameters used in our WIST training. Unless otherwise specified, we set the target domain to \textbf{Mathematics}. In addition, we use a Sentence Embedding model, namely all-MiniLM-L6-v2~\cite{reimers-2019-sentence-bert}, as our semantic encoder $\phi(\cdot)$.

\begin{table}[t]
\centering
\small
\setlength{\tabcolsep}{6pt}
\renewcommand{\arraystretch}{1.1}
\begin{tabular}{l l}
\toprule
\textbf{Hyperparameter} & \textbf{Value} \\
\midrule
Rollout batch size (paths per iteration) $B$ & 128 \\
Training batch size $B_{\mathrm{train}}$ & 512 \\
Tree maximum depth $\mathcal{L}$ & 4 \\
Beta prior $(\alpha_0,\beta_0)$ & $(1,1)$ \\
Learnability threshold (variance) $\tau_{\mathrm{var}}$ & 0.20 \\
Sliding window size $\mu$ & 5 \\
Title semantic similarity threshold $\tau_{\mathrm{emb}}$ & 0.5 \\
Wiki validation threshold $\tau_{\mathrm{wiki}}$ & 0.8 \\
Challenger QA candidates $G$ & 8 \\
Solver self-consistency samples $K$ & 8 \\
Rule-validation penalty $\rho$ & $-0.1$ \\
Challenger shaping width $\sigma$ & 0.02 \\
KL coefficient $\lambda_{\mathrm{KL}}$ & 0.0 \\
Learning rate $\eta$ & $1\times 10^{-6}$ \\
Total training updates $T_{\mathrm{steps}}$ & 50 \\
\bottomrule
\end{tabular}
\caption{Main hyperparameters for WIST.}
\label{tab:wist_hparams}
\end{table}

\section{Evaluation Settings}
\label{sec:evaluation_settings}
\paragraph{Evaluation Protocol.}
We evaluate all models in a \emph{zero-shot} setting to examine whether the reasoning abilities acquired through corpus-grounded self-play generalize to standard benchmarks without any task-specific adaptation. For most benchmarks, we use \emph{greedy decoding} (temperature $=0$) to maximize reproducibility and ensure consistent comparisons across models.

\paragraph{Mathematical Reasoning.}
For \textsc{AIME'24} and \textsc{AIME'25}, we adopt a sampling-based protocol to better capture performance on challenging competition-style problems: we run $32$ independent generations with temperature $=0.6$ and report the \emph{average accuracy}. For the remaining mathematical reasoning benchmarks, including \textsc{MATH-500}, \textsc{OlympiadBench}, \textsc{Minerva Math}, \textsc{GSM8K}, and \textsc{AMC}, we report \emph{pass@1} under greedy decoding. Predictions are scored by \emph{exact match} after answer extraction and normalization. To reduce false negatives caused by formatting differences, we additionally perform equivalence checking via \textsc{gpt-4.1-2025-04-14} verification.

\paragraph{General Reasoning.}
For general reasoning, we evaluate on \textsc{GPQA-Diamond}, \textsc{SuperGPQA}, \textsc{MMLUPro}, and \textsc{BBEH}. All general-reasoning evaluations use greedy decoding and are scored by \emph{exact match} on the extracted multiple-choice option (A/B/C/D). We keep prompts consistent across models by using the same system prompt and answer extraction format as in training. Evaluation prompts instruct models to produce step-by-step reasoning before emitting a final answer, formatted as a boxed result for mathematical tasks or as a letter choice for multiple-choice questions. We will release evaluation prompts and code to support reproducibility.


\section{Additional Results and Analysis}
\label{sec:additional_results}

This section provides complementary results and analyses that further characterize WIST. 
We first report full-benchmark comparisons between two exploration variants (random vs.\ reward). 
We then study cross-model transfer of the learned \emph{domain tree} by constructing the tree with a stronger model and training a smaller model with the transferred tree. 
Finally, we demonstrate that WIST can be steered to a scientific domain beyond math/physics, namely medicine, and evaluate on three medical QA benchmarks.

\subsection{Additional Scaling Results on larger model}
\label{sec:qwen14b_appendix}
In the main paper, WIST is slightly below SPICE in the small-model setting (\textit{OctoThinker-3B-Hybrid-Base}). We attribute this mainly to two factors: (1) smaller models are less reliable in domain-tree expansion and thus more prone to noisy or less informative sub-concepts; and (2) smaller models are more sensitive to residual noise in open-web data, whereas SPICE benefits from a cleaner curated environment.

To further examine the effect of model scale, we additionally evaluate WIST on \texttt{Qwen3-14B-Base}. Table~\ref{tab:qwen3_14b_appendix} reports the full results on the same set of mathematical and general reasoning benchmarks used in the main paper. We observe that WIST achieves the best overall performance, improving over base model by \textbf{+3.2} on the Overall score. This result supports the view that tree-guided open-web grounding becomes more effective as the capability of the base model increases.

\begin{table*}[t]
	\centering
	\footnotesize
	\setlength{\tabcolsep}{1.2pt}
	\renewcommand{\arraystretch}{1.1}
	\caption{\textbf{Additional results on \texttt{Qwen3-14B-Base}} across mathematical and general reasoning benchmarks. Best and second-best results within this backbone block are marked in \textbf{bold} and \underline{underline}, respectively.}
	\label{tab:qwen3_14b_appendix}
	\resizebox{0.95\textwidth}{!}{%
		\begin{tabular}{l|ccccccc|cccc|c}
			\toprule
			& \multicolumn{7}{c|}{\textbf{Mathematical Reasoning}} & \multicolumn{4}{c|}{\textbf{General Reasoning}} & \\
			\cmidrule(lr){2-8}\cmidrule(lr){9-12}
			\textbf{Method} &
			\textbf{AMC} & \textbf{Minerva} & \makecell{\textbf{MATH}\\\textbf{500}} & \textbf{GSM8K} & \textbf{Olymp.} &
			\makecell{\textbf{AIME}\\\textbf{24}} & \makecell{\textbf{AIME}\\\textbf{25}} &
			\makecell{\textbf{Super-}\\\textbf{GPQA}} & \makecell{\textbf{GPQA-}\\\textbf{Diamond}} & \makecell{\textbf{MMLU-}\\\textbf{Pro}} & \textbf{BBEH} &
			\textbf{Overall} \\
			\midrule
			
			\rowcolor{gray!10}\multicolumn{13}{c}{\textit{Qwen3-14B-Base}}\\
			Base Model                     & \underline{62.5} & 50.0 & 80.2 & 93.7 & 43.1 & 13.3 & 11.4 & 35.2 & 40.4 & 63.9 & 14.0 & 46.2 \\
			\hspace{1em}+ R-Zero           & 60.3 & 52.9 & 81.4 & 94.4 & 45.2 & 14.3 & 11.8 & \underline{37.1} & \underline{44.4} & 65.3 & 14.9 & 47.5 \\
			\hspace{1em}+ SPICE            & 60.2 & \underline{56.3} & \textbf{83.8} & \underline{94.5} & \underline{46.7} & \underline{15.2} & \underline{15.1} & \underline{37.1} & \textbf{45.5} & \underline{65.4} & \underline{15.2} & \underline{48.6} \\
			\rowcolor{blue!10}
			\textbf{\hspace{1em}+ WIST (ours)} & \textbf{67.5} & \textbf{56.6} & \underline{83.4} & \textbf{94.7} & \textbf{46.8} & \textbf{15.4} & \textbf{15.8} & \textbf{37.4} & \underline{44.4} & \textbf{66.0} & \textbf{15.3} & \textbf{49.4} \\
			\bottomrule
		\end{tabular}
	}
\end{table*}

\subsection{Random vs.\ Reward Tree Expansion}  
\label{sec:random_reward}
Table~\ref{tab:random_reward_full} reports the results of all benchmark tests under the same training budget and evaluation protocol for the two variants of WIST, i.e. random and reward.

\begin{table*}[t]
	\centering
	\footnotesize
	\setlength{\tabcolsep}{1.2pt}
	\renewcommand{\arraystretch}{1.0}
	\caption{Full-benchmark comparison between random and reward-guided tree expansion under the same training budget.}
	\resizebox{0.95\textwidth}{!}{%
		\begin{tabular}{l|ccccccc|cccc|c}
			\toprule
			& \multicolumn{7}{c|}{\textbf{Mathematical Reasoning}} & \multicolumn{4}{c|}{\textbf{General Reasoning}} & \\
			\cmidrule(lr){2-8}\cmidrule(lr){9-12}
			\textbf{Variant} &
			\textbf{AMC} & \textbf{Minerva} & \makecell{\textbf{MATH}\\\textbf{500}} & \textbf{GSM8K} & \textbf{Olymp.} &
			\makecell{\textbf{AIME}\\\textbf{24}} & \makecell{\textbf{AIME}\\\textbf{25}} &
			\makecell{\textbf{Super-}\\\textbf{GPQA}} & \makecell{\textbf{GPQA-}\\\textbf{Diamond}} & \makecell{\textbf{MMLU-}\\\textbf{Pro}} & \textbf{BBEH} &
			\textbf{Overall} \\
			\midrule
			\rowcolor{gray!10}\multicolumn{13}{c}{\textit{Qwen3-4B-Base}}\\
			Base Model                     & 41.4 & 35.7 & 57.0 & 75.9 & 30.2 & 9.5 &  6.4 & 18.0 & 32.8 & 51.5 &  8.2 & 33.3 \\
			\hspace{1em}+ WIST (Random)         & \underline{51.1} & \underline{46.0} & \underline{78.0} & \underline{92.8} & \textbf{40.1} & \textbf{11.9} & \textbf{9.8} & \underline{28.0} & \textbf{37.9} & \underline{53.5} & \underline{11.7} & \underline{41.9} \\
			\rowcolor{blue!10}
			\hspace{1em}+ \textbf{WIST (Reward)}        & \textbf{60.0} & \textbf{47.8} & \textbf{78.2} & \textbf{92.9} & \underline{40.0} & \underline{11.6} & \underline{9.7} & \textbf{29.6} & \underline{37.2} & \textbf{55.7} & \textbf{11.8} & \textbf{43.1} \\
			\bottomrule
		\end{tabular}
	}
	\label{tab:random_reward_full}
\end{table*}

\subsection{Targeted Domain Steering to Medicine}
\label{sec:medical_results}
We report in Table~\ref{tab:medical_domain_results} the detailed benchmark results of the best-performing checkpoint obtained during medical-domain training.

\begin{table}[t]
	\centering
	\footnotesize
	\setlength{\tabcolsep}{3.5pt}
	\renewcommand{\arraystretch}{1.1}
	\caption{\textbf{Medical-domain results} after steering the target domain to medicine. We report benchmark-level accuracy on Medbullets, MedMCQA, and MedQA, together with their average. Best and second-best results within each backbone block can be marked in \textbf{bold} and \underline{underline}, respectively.}
	\label{tab:medical_domain_results}
	\resizebox{1.0\columnwidth}{!}{%
		\begin{tabular}{l|ccc|c}
			\toprule
			\textbf{Method} & \textbf{Medbullets} & \textbf{MedMCQA} & \textbf{MedQA} & \textbf{Avg.} \\
			\midrule
			
			\rowcolor{gray!10}\multicolumn{5}{c}{\textit{Qwen3-4B-Base}}\\
			Base Model                       & 24.03 & 32.27 & 48.52 & 34.94 \\
			\hspace{1em}+ R-Zero             & 24.03 & 32.27 & 51.50 & 35.93 \\
			\hspace{1em}+ SPICE              & \textbf{24.35} & \textbf{32.39} & \underline{52.69} & \underline{36.48} \\
			\hspace{1em}+ WIST (Random)      & 24.19 & 32.30 & 52.76 & 36.42 \\
			\rowcolor{blue!10}
			\textbf{\hspace{1em}+ WIST (Reward)} & \textbf{24.35} & \underline{32.37} & \textbf{53.42} & \textbf{36.71} \\
			\midrule
			
			\rowcolor{gray!10}\multicolumn{5}{c}{\textit{Qwen3-8B-Base}}\\
			Base Model                       & 30.03 & 39.80 & 50.87 & 40.24 \\
			\hspace{1em}+ R-Zero             & 34.58 & 46.07 & 54.54 & 45.06 \\
			\hspace{1em}+ SPICE              & 45.94 & 54.36 & \textbf{58.94} & 53.08 \\
			\hspace{1em}+ WIST (Random)      & \textbf{50.16} & \underline{55.63} & 57.22 & \underline{54.34} \\
			\rowcolor{blue!10}
			\textbf{\hspace{1em}+ WIST (Reward)} & \underline{50.00} & \textbf{56.35} & \underline{58.73} & \textbf{55.03} \\
			
			\midrule
			\rowcolor{gray!10}\multicolumn{5}{c}{\textit{Qwen3-14B-Base}}\\
			Base Model                       & 38.64 & 48.51 & 60.15 & 49.10 \\
			\hspace{1em}+ R-Zero             & 46.43 & 51.83 & 61.48 & 53.25 \\
			\hspace{1em}+ SPICE              & \underline{50.16} & \underline{56.87} & 61.96 & \underline{56.33} \\
			\hspace{1em}+ WIST (Random)      & 50.00 & 53.96 & \underline{62.30} & 55.42 \\
			\rowcolor{blue!10}
			\textbf{\hspace{1em}+ WIST (Reward)} & \textbf{54.22} & \textbf{60.29} & \textbf{62.66} & \textbf{59.06} \\
			
			\bottomrule
		\end{tabular}
	}
\end{table}

\subsection{Cross-model Tree Transfer: Strong Builder, Small Learner}
\label{sec:tree_transfer}
A practical advantage of WIST is that the domain tree is an explicit, reusable artifact.
We study whether a high-quality tree built by a stronger model can be transferred to improve the training of a smaller model.
Specifically, we use \textit{Qwen3-14B} to expand a mathematics domain tree (including leaf-level atomic knowledge points) under the same web filtering and validation rules, and then freeze this tree for subsequent training of \textit{Qwen3-4B-Base}.
We compare against (1) a tree built by the small model itself (Self-Tree), and (2) a tree build by the strong model (Strong-Tree), all under matched training steps and rollout budget.

\begin{table*}[t]
	\centering
	\footnotesize
	\setlength{\tabcolsep}{1.2pt}
	\renewcommand{\arraystretch}{1.0}
	\caption{Tree transfer from a stronger builder model (\textit{Qwen3-14B}) to a smaller learner (\textit{Qwen3-4B-Base}) in mathematics.}
	\resizebox{0.95\textwidth}{!}{%
		\begin{tabular}{l|ccccccc|cccc|c}
			\toprule
			& \multicolumn{7}{c|}{\textbf{Mathematical Reasoning}} & \multicolumn{4}{c|}{\textbf{General Reasoning}} & \\
			\cmidrule(lr){2-8}\cmidrule(lr){9-12}
			\textbf{Variant} &
			\textbf{AMC} & \textbf{Minerva} & \makecell{\textbf{MATH}\\\textbf{500}} & \textbf{GSM8K} & \textbf{Olymp.} &
			\makecell{\textbf{AIME}\\\textbf{24}} & \makecell{\textbf{AIME}\\\textbf{25}} &
			\makecell{\textbf{Super-}\\\textbf{GPQA}} & \makecell{\textbf{GPQA-}\\\textbf{Diamond}} & \makecell{\textbf{MMLU-}\\\textbf{Pro}} & \textbf{BBEH} &
			\textbf{Avg.} \\
			\midrule
			Base Model                     & 41.4 & 35.7 & 57.0 & 75.9 & 30.2 & 9.5 &  6.4 & 18.0 & 32.8 & 51.5 &  8.2 & 33.3 \\
			\hspace{1em}+ WIST (Strong-Tree)         & 53.2 & 49.3 & 78.2 & 92.5 & 39.7 & 11.5 & 9.6 & 28.5 & 37.4 & 55.4 & 11.7 & 42.5 \\
			\hspace{1em}+ WIST (Self-Tree)        & 60.0 & 47.8 & 78.2 & 92.9 & 40.0 & 11.6 & 9.7 & 29.6 & 37.2 & 55.7 & 11.8 & 43.1 \\
			\bottomrule
		\end{tabular}
	}
	\label{tab:strong}
\end{table*}

From Table~\ref{tab:strong}, Self-Tree and Strong-Tree both substantially improve over the base model, but \textbf{Strong-Tree} remains slightly behind \textbf{Self-Tree}. This indicates that although a stronger model can construct a reasonable and reusable tree, \emph{tree quality alone} does not necessarily translate into better training outcomes for a weaker learner.

We attribute this behavior to two factors:
\begin{itemize}
    \item \textbf{Learner--curriculum mismatch.}: The tree expanded by \textit{Qwen3-14B} tends to introduce finer-grained and harder leaf concepts, which more often yield extreme self-consistency signals for \textit{Qwen3-4B-Base} and thus produce fewer learnable self-play instances. In contrast, a self-built tree is better aligned with the learner's capability boundary.
    \item \textbf{Posterior dynamics and guidance quality.}: Since WIST relies on posterior-guided exploration, differences in the tree structure can change how quickly and stably node posteriors concentrate, affecting the strength of the guidance signal. Self-Tree typically yields more stable guidance for the learner, resulting in slightly better final performance.
\end{itemize}

\subsection{Open-Web Data Processing and Ablation}

\label{sec:web_data_processing}

Using the open web as a training environment introduces both opportunity and risk: while it provides abundant and continuously refreshed domain knowledge, it also contains noise, redundancy, and potential benchmark contamination. To mitigate these issues, WIST performs a multi-stage data processing pipeline before self-play. Our design is inspired by web-scale cleaning and curation practices in \textbf{NVIDIA NeMo Curator}, while being adapted to our tree-guided retrieval setting.

\subsubsection{Data Processing Pipeline}
\label{sec:data_processing_pipeline}

Given a sampled leaf concept from the domain tree, WIST retrieves candidate web pages and processes them before adding them to the corresponding corpus pool. The pipeline consists of the following steps:

\begin{enumerate}
	\item \textbf{URL and source filtering.} We first filter retrieved URLs using allow/deny rules to exclude low-quality or undesirable sources. This step is used to reduce obvious noise and lower the risk of benchmark leakage.
	
	\item \textbf{Concept--page relevance filtering.} We then measure the semantic relevance between the target concept and the retrieved page (e.g., using the page title and concept representation). Only pages whose relevance exceeds a threshold are kept, ensuring that the resulting corpus remains aligned with the sampled tree node.
	
	\item \textbf{Boilerplate removal.} For each retained page, we remove non-content HTML components such as navigation bars, advertisements, templates, and repeated layout blocks. This step extracts the main textual content that is more suitable for subsequent self-play.
	
	\item \textbf{Normalization and deduplication.} Finally, we normalize the cleaned text and remove near-duplicate content to reduce redundancy across retrieved pages and improve the stability of the resulting corpus pool.
\end{enumerate}

Together, these steps convert noisy open-web retrieval results into a more controllable and leaf-aligned corpus environment for self-play training.

\subsubsection{Ablation on Data Processing}
\label{sec:data_processing_ablation}

To quantify the effect of data processing, we compare our default \textbf{full filtering} pipeline against a weaker variant with \textbf{HTML-only} cleaning. The latter performs only basic HTML content extraction / cleaning, without the URL/source filtering,  relevance filtering, normalization and deduplication.

\begin{table*}[t]
	\centering
	\footnotesize              
	\setlength{\tabcolsep}{1.2pt}  
	\renewcommand{\arraystretch}{1.1} 
	\caption{\textbf{Ablation on web data processing.} We report the performance over all mathematical and general reasoning benchmarks.}
	\label{tab:web_data_processing_ablation}
	\resizebox{0.95\textwidth}{!}{%
		\begin{tabular}{l|ccccccc|cccc|c}
			\toprule
			& \multicolumn{7}{c|}{\textbf{Mathematical Reasoning}} & \multicolumn{4}{c|}{\textbf{General Reasoning}} & \\
			\cmidrule(lr){2-8}\cmidrule(lr){9-12}
			\textbf{Method} &
			\textbf{AMC} & \textbf{Minerva} & \makecell{\textbf{MATH}\\\textbf{500}} & \textbf{GSM8K} & \textbf{Olymp.} &
			\makecell{\textbf{AIME}\\\textbf{24}} & \makecell{\textbf{AIME}\\\textbf{25}} &
			\makecell{\textbf{Super-}\\\textbf{GPQA}} & \makecell{\textbf{GPQA-}\\\textbf{Diamond}} & \makecell{\textbf{MMLU-}\\\textbf{Pro}} & \textbf{BBEH} &
			\textbf{Overall} \\
			\midrule
			
			\rowcolor{gray!10}\multicolumn{13}{c}{\textit{Qwen3-4B-Base}}\\
			Base Model                     & 41.4 & 35.7 & 57.0 & 75.9 & 30.2 & 9.5 &  6.4 & 18.0 & 32.8 & 51.5 &  8.2 & 33.3 \\
			\hspace{1em}+ WIST (HTML-only)       & 56.6 & 48.6 & 77.0 & 91.8 & 40.0 & 11.3 & 9.5 & 28.6 & 31.8 & 55.1 & 12.1 & 42.0 \\
			\hspace{1em}+ WIST (full filtering)  & 60.0 & 47.8 & 78.2 & 92.9 & 40.0 & 11.6 & 9.7 & 29.6 & 37.2 & 55.7 & 11.8 & 43.1 \\
			\midrule
			
			\rowcolor{gray!10}\multicolumn{13}{c}{\textit{Qwen3-8B-Base}}\\
			Base Model                     & 57.2 & 43.0 & 73.0 & 91.3 & 40.5 & 11.7 & 11.3 & 28.3 & 34.8 & 58.2 & 9.1 & 42.1 \\
			\hspace{1em}+ WIST (HTML-only)       & 59.3 & 51.5 & 81.6 & 93.0 & 44.4 & 15.3 & 13.8 & 31.2 & 40.9 & 61.1 & 13.4 & 46.0 \\
			\hspace{1em}+ WIST (full filtering)  & 63.4 & 53.3 & 82.6 & 93.4 & 44.1 & 14.8 & 13.9 & 32.5 & 41.4 & 61.6 & 12.9 & 46.7 \\
			\midrule\midrule
			
			\rowcolor{gray!10}\multicolumn{13}{c}{\textit{OctoThinker-3B-Hybrid-Base}}\\
			Base Model                    & 12.5&18.5&30.6&44.9&11.0&1.7&\underline{0.6}&10.4&2.0&11.1&2.3&13.2 \\
			\hspace{1em}+ WIST (HTML-only)       & 23.6 & 21.0 & 49.6 & 74.4 & 16.7 & 1.9 & 0.6 & 17.4 & 23.2 & 31.7 & 4.2 & 24.0 \\
			\hspace{1em}+ WIST (full filtering)  & 27.4 & 22.7 & 48.8 & 76.3 & 15.1 & 1.9 & 0.6 & 17.8 & 24.1 & 30.4 & 4.1 & 24.5 \\
			\midrule
			
			\rowcolor{gray!10}\multicolumn{13}{c}{\textit{OctoThinker-8B-Hybrid-Base}}\\
			Base Model                     & 20.0&26.2&42.8&82.2&17.0&2.4&1.1&16.4&12.1&25.9&5.4&22.9 \\
			\hspace{1em}+ WIST (HTML-only)       & 30.7 & 35.3 & 58.0 & 85.7 & 21.9 & 5.0 & 1.4 & 22.7 & 28.8 & 40.1 & 10.2 & 30.9 \\
			\hspace{1em}+ WIST (full filtering)  & 31.0 & 36.4 & 62.0 & 87.0 & 25.5 & 3.2 & 1.4 & 25.4 & 30.6 & 45.9 & 10.1 & 32.6 \\
			\bottomrule
		\end{tabular}
	}
\end{table*}

As shown in Table~\ref{tab:web_data_processing_ablation}, full filtering and cleaning yields consistent gains across all backbones, improving the mean score by \textbf{+0.5} to \textbf{+1.7} points over HTML-only cleaning. This indicates that careful web-data curation is important for stable and reliable open-web learning. In particular, basic HTML cleaning alone is not sufficient: URL/source filtering, concept relevance filtering, and deduplication all contribute to building a higher-quality self-play environment.

 \section{Prompt Templates}
 \label{sec:prompt_templates}

 \subsection{Prompt Template for Node Expansion}  

 We present the prompt templates used for expanding the domain tree. Depending on whether the sampled \textit{unk} node appears at a non-leaf level or at the leaf level, the model is prompted to propose either (i) a new \emph{sub-domain} or (ii) a new \emph{atomic knowledge point}. Both templates share a unified ``No More'' convention and a strict response format to reduce non-standard proposals and stabilize downstream parsing. \\
 \textbf{Placeholders.} \textit{\{main\_domain\}} is the user-specified target domain (e.g., Mathematics/Physics), \textit{\{path\_in\_str\}} is the current root-to-node path, \textit{\{parent\_name\}} is the parent node to be expanded, \textit{\{existing\_children\}} lists existing children under \textit{\{parent\_name\}}, and \textit{\{siblings\}} optionally lists sibling domains for lightweight anti-misattachment guidance.

 \begin{figure*}[t]
 	\centering
 	\begin{tcolorbox}[
 		enhanced,
 		breakable,
 		colback=white,
 		colframe=blue!50!red,
 		colbacktitle=blue!50!red,
 		coltitle=white,
 		title={Sub-domain Expansion Prompt (Non-leaf Nodes)},
 		center title,
 		fonttitle=\bfseries,
 		boxrule=1.2pt,
 		arc=1.8mm,
 		left=1.2mm,
 		right=1.2mm,
 		top=1.0mm,
 		bottom=0.8mm,
 		boxsep=0.8mm
 		]
 		\footnotesize
 		
 		You are helping to construct a hierarchical knowledge tree in the main domain:
 		\texttt{\{main\_domain\}}.
 		The current node path is:
 		\texttt{\{path\_in\_str\}}.
 		
 		\vspace{0.25em}
 		\textbf{Your task: }
 		Propose \textbf{ONE NEW SUB-DOMAIN} that will become a direct child of
 		\texttt{"\{parent\_name\}"}.
 		
 		\vspace{0.25em}
 		This tree is intended for school and early-university level \texttt{\{main\_domain\}}. It will be used to generate exam-style and competition-style problems (multiple-choice or short-answer), similar to AMC / AIME / olympiad-style, as well as standard early undergraduate courses.
 		
 		\vspace{0.25em}
 		\textbf{Optional sibling guidance} (include only if siblings are provided).
 		
 		\textbf{Context about siblings} (soft guidance):
 		\begin{itemize}[leftmargin=1.4em,itemsep=0.15em,topsep=0.15em,parsep=0pt,partopsep=0pt]
 			\item Under the same parent as \texttt{"\{parent\_name\}"}, there are already some sibling sub-domains:
 			\texttt{\{siblings\}}.
 			\item This list is provided \textbf{only} to help you avoid:
 			\begin{itemize}[leftmargin=1.4em,itemsep=0.1em,topsep=0.1em,parsep=0pt,partopsep=0pt]
 				\item proposing a label that is almost the same as a sibling, or
 				\item proposing a topic that is obviously a subtopic of a sibling instead of \texttt{"\{parent\_name\}"}.
 			\end{itemize}
 			\item The primary objective is still to create a standard curriculum-style sub-domain for
 			\texttt{"\{parent\_name\}"}.
 		\end{itemize}
 		
 		\vspace{0.25em}
 		\textbf{Illustrative examples of good hierarchical structure} (examples only):
 		\begin{itemize}[leftmargin=1.4em,itemsep=0.15em,topsep=0.15em,parsep=0pt,partopsep=0pt]
 			\item \texttt{\{main\_domain\}} $\rightarrow$ Algebra / Geometry / Number Theory / \ldots
 			\item Typical sub-domains under \texttt{"Algebra"}:
 			Equations and Inequalities / Polynomials / \ldots
 		\end{itemize}
 		
 		\vspace{0.25em}
 		\textbf{Strict requirements for the new sub-domain.}
 		
 		\begin{enumerate}[leftmargin=1.5em,itemsep=0.2em,topsep=0.2em,parsep=0pt,partopsep=0pt]
 			\item \textbf{Subset relation and level of generality.}
 			\begin{itemize}[leftmargin=1.4em,itemsep=0.1em,topsep=0.1em,parsep=0pt,partopsep=0pt]
 				\item The new sub-domain must be strictly more specific than \texttt{"\{parent\_name\}"}.
 				\item It should look like a chapter/section title in a school/early-university textbook.
 			\end{itemize}
 			
 			\item \textbf{Difficulty and scope constraint (primary).}
 			\begin{itemize}[leftmargin=1.4em,itemsep=0.1em,topsep=0.1em,parsep=0pt,partopsep=0pt]
 				\item Focus on standard curriculum topics; avoid graduate-level or research-only topics.
 			\end{itemize}
 			
 			\item \textbf{Existing children under this parent (local de-duplication).}
 			\begin{itemize}[leftmargin=1.4em,itemsep=0.1em,topsep=0.1em,parsep=0pt,partopsep=0pt]
 				\item The new sub-domain must \textbf{not} be identical or almost identical to any existing child:
 				\texttt{\{existing\_children\}}.
 			\end{itemize}
 			
 			\item \textbf{Naming style.}
 			\begin{itemize}[leftmargin=1.4em,itemsep=0.1em,topsep=0.1em,parsep=0pt,partopsep=0pt]
 				\item Use clear and concise names; avoid unnatural over-specific phrasing.
 			\end{itemize}
 			
 			\item \textbf{Anti-hallucination rule (crucial).}
 			\begin{itemize}[leftmargin=1.4em,itemsep=0.1em,topsep=0.1em,parsep=0pt,partopsep=0pt]
 				\item You \textbf{must not} invent new theorem names or dubious terminology.
 				\item If unsure, choose a simpler, classical topic instead.
 			\end{itemize}
 			
 			\item \textbf{Domain purity.}
 			\begin{itemize}[leftmargin=1.4em,itemsep=0.1em,topsep=0.1em,parsep=0pt,partopsep=0pt]
 				\item The label must stay within the target domain and should not reference other domains.
 			\end{itemize}
 		\end{enumerate}
 		
 		\vspace{0.25em}
 		If you believe there are \textbf{no further meaningful sub-domains} under
 		\texttt{"\{parent\_name\}"} that:
 		\begin{itemize}[leftmargin=1.4em,itemsep=0.15em,topsep=0.15em,parsep=0pt,partopsep=0pt]
 			\item are not already covered by the existing children listed above,
 			\item fit the school / early-university scope, and
 			\item correspond to standard, widely used topics,
 		\end{itemize}
 		then you must output exactly \texttt{"No More"}.
 		
 		\vspace{0.25em}
 		\textbf{Strict response format.}
 		\begin{itemize}[leftmargin=1.4em,itemsep=0.15em,topsep=0.15em,parsep=0pt,partopsep=0pt]
 			\item Propose \textbf{exactly one} new label.
 			\item Enclose it between \texttt{[Proposition Start]} and \texttt{[Proposition End]}, e.g.:
 		\end{itemize}
 		
 		\begin{center}
 			\texttt{[Proposition Start]Quadratic Equations[Proposition End]}
 		\end{center}
 		
 		Now, provide your proposed label.
 		
 	\end{tcolorbox}
 \end{figure*}

 
  \begin{figure*}[t]
 	\centering
 	\begin{tcolorbox}[
 		enhanced,
		breakable,
		colback=white,
		colframe=blue!50!red,
		colbacktitle=blue!50!red,
		coltitle=white,
		title={Knowledge-Point Expansion Prompt (Leaf Nodes)},
		center title,
		fonttitle=\bfseries,
		boxrule=1.2pt,
		arc=1.8mm,
		left=1.2mm,
		right=1.2mm,
		top=1.0mm,
		bottom=0.8mm,
		boxsep=0.8mm
 		]
 		\footnotesize
 		
 		You are helping to construct a hierarchical knowledge tree in the main domain:
 		\texttt{\{main\_domain\}}.
 		The current node path is:
 		\texttt{\{path\_in\_str\}}.
 		
 		\textbf{Your task:} propose \textbf{ONE NEW ATOMIC KNOWLEDGE POINT} that will become a direct child of
 		\texttt{"\{parent\_name\}"}.
 		
 		A knowledge point must be a minimal unit that is directly usable to construct exam/contest-style problems,
 		such as a named theorem/lemma/proposition, a standard definition used in problems, a classical example,
 		or a standard algorithm/construction.
 		
 		\texttt{\{Optional sibling guidance (lightweight, include only if siblings are provided):\}}
 		
 		\textbf{Sibling knowledge points (soft guidance):}
 		\begin{itemize}[leftmargin=1.4em,itemsep=0.15em,topsep=0.15em,parsep=0pt,partopsep=0pt]
 			\item Under the same parent \texttt{"\{parent\_name\}"}, there may already be other knowledge points:
 			\texttt{\{existing\_children\}}
 			\item Avoid near-duplicates; siblings are only a local de-duplication hint.
 		\end{itemize}
 		
 		\textbf{STRICT REQUIREMENTS FOR THE NEW KNOWLEDGE POINT:}
 		
 		\begin{enumerate}[leftmargin=1.5em,itemsep=0.2em,topsep=0.2em,parsep=0pt,partopsep=0pt]
 			\item \textbf{Scope and granularity}
 			\begin{itemize}[leftmargin=1.4em,itemsep=0.1em,topsep=0.1em,parsep=0pt,partopsep=0pt]
 				\item It must be strictly narrower than \texttt{"\{parent\_name\}"} and correspond to \textbf{EXACTLY ONE} atomic unit:
 				\begin{itemize}[leftmargin=1.4em,itemsep=0.1em,topsep=0.1em,parsep=0pt,partopsep=0pt]
 					\item theorem / lemma / proposition, OR
 					\item standard definition, OR
 					\item classical configuration/example, OR
 					\item standard algorithm/construction.
 				\end{itemize}
 				\item It should look like a short standalone textbook entry.
 			\end{itemize}
 			
 			\item \textbf{Difficulty and usability (primary)}
 			\begin{itemize}[leftmargin=1.4em,itemsep=0.1em,topsep=0.1em,parsep=0pt,partopsep=0pt]
 				\item It should support multi-step but standard contest / early-undergrad problems.
 				\item Avoid research-level, overly advanced, or non-canonical topics.
 			\end{itemize}
 			
 			\item \textbf{Existing children under this parent (local de-duplication)}
 			\begin{itemize}[leftmargin=1.4em,itemsep=0.1em,topsep=0.1em,parsep=0pt,partopsep=0pt]
 				\item The new knowledge point must \textbf{NOT} be identical or almost identical to any existing child:
 				\texttt{\{existing\_children\}}
 			\end{itemize}
 			
 			\item \textbf{Naming style}
 			\begin{itemize}[leftmargin=1.4em,itemsep=0.1em,topsep=0.1em,parsep=0pt,partopsep=0pt]
 				\item Use clear and concise names; do \textbf{NOT} always choose \texttt{"Definition: ..."};
 				mix theorems/examples/algorithms when natural.
 			\end{itemize}
 			
 			\item \textbf{Anti-hallucination rule (crucial)}
 			\begin{itemize}[leftmargin=1.4em,itemsep=0.1em,topsep=0.1em,parsep=0pt,partopsep=0pt]
 				\item You \textbf{MUST NOT} invent new theorem names, lemma names, or terminology.
 				\item Only propose names that are standard and widely used in textbooks.
 			\end{itemize}
 			
 			\item \textbf{Domain purity}
 			\begin{itemize}[leftmargin=1.4em,itemsep=0.1em,topsep=0.1em,parsep=0pt,partopsep=0pt]
 				\item The name must stay purely in the target domain.
 			\end{itemize}
 		\end{enumerate}
 		
 		If you believe there are \textbf{NO} further meaningful knowledge points under
 		\texttt{"\{parent\_name\}"} that:
 		\begin{itemize}[leftmargin=1.4em,itemsep=0.15em,topsep=0.15em,parsep=0pt,partopsep=0pt]
 			\item are not already covered by the existing children listed above, \textbf{AND}
 			\item fit the intended scope, \textbf{AND}
 			\item correspond to standard, widely used topics,
 		\end{itemize}
 		then you must output exactly \texttt{"No More"}.
 		
 		\textbf{STRICT RESPONSE FORMAT:}
 		\begin{itemize}[leftmargin=1.4em,itemsep=0.15em,topsep=0.15em,parsep=0pt,partopsep=0pt]
 			\item Propose \textbf{EXACTLY ONE} new label.
 			\item Enclose it between \texttt{[Proposition Start]} and \texttt{[Proposition End]}, e.g.:
 		\end{itemize}
 		
 		\begin{center}
 			\texttt{[Proposition Start]Pigeonhole Principle[Proposition End]}
 		\end{center}
 		
 		Now, provide your proposed label.
 		
 	\end{tcolorbox}
 \end{figure*}

   \begin{figure*}[t]
   	\vspace{-10mm}
 	\centering
 	\begin{tcolorbox}[
 		enhanced,
 		breakable,
 		colback=white,
 		colframe=blue!50!red,
 		colbacktitle=blue!50!red,
 		coltitle=white,
 		title={MCQ Question Generation Prompt (Path-Conditioned, adapted from SPICE)},
 		center title,
 		fonttitle=\bfseries,
 		boxrule=1.2pt,
 		arc=1.8mm,
 		left=1.2mm,
 		right=1.2mm,
 		top=0.4mm,
 		bottom=0.3mm,
 		boxsep=0.3mm
 		]
 		\footnotesize
 		
 		Your task is to create a \textbf{CHALLENGING} question from a document by using \textbf{BOTH}:
 		\begin{itemize}[leftmargin=1.4em,itemsep=0.1em,topsep=0.1em,parsep=0pt,partopsep=0pt]
 			\item (1) a hierarchical \textbf{LABEL PATH} that narrows down the mathematical domain, and
 			\item (2) background \textbf{TEXT} about the most specific knowledge point.
 		\end{itemize}
 		
 		\textbf{\#\# Label Path (Domain Hierarchy)}\\
 		\texttt{[BEGINNING OF THE LABEL PATH]}\\
 		\texttt{\{path\}}\\
 		\texttt{[END OF THE LABEL PATH]}
 		
 		The label path lists nested mathematical domains from the broadest on the left to the most specific on the right.\\
 		Example: \texttt{"Mathematics -> Algebra -> Group Theory -> Sylow's Theorems"}
 		
 		\begin{itemize}[leftmargin=1.4em,itemsep=0.1em,topsep=0.1em,parsep=0pt,partopsep=0pt]
 			\item The \textbf{LEFTMOST} labels are broad fields (e.g., \texttt{"Mathematics"}, \texttt{"Algebra"}).
 			\item The \textbf{RIGHTMOST} label is an \textbf{ATOMIC KNOWLEDGE POINT} (e.g., \texttt{"Sylow's Theorems"}).
 			\item Your question \textbf{MUST} belong to this path:
 			\begin{itemize}[leftmargin=1.4em,itemsep=0.1em,topsep=0.1em,parsep=0pt,partopsep=0pt]
 				\item It must clearly be a mathematics question.
 				\item It must primarily test the \textbf{RIGHTMOST} knowledge point.
 				\item It may use context from earlier levels in the path to add difficulty and require multi-step reasoning.
 			\end{itemize}
 			\item If the text contains information that is irrelevant to this label path, \textbf{IGNORE} that information.
 		\end{itemize}
 		
 		\textbf{\#\# Text}\\
 		\texttt{[BEGINNING OF THE DOCUMENT]}\\
 		\texttt{\{text\}}\\
 		\texttt{[END OF THE DOCUMENT]}
 		
 		The text is background material (e.g., web pages) about the atomic knowledge point at the end of the label path.\\
 		You must use this text to construct a mathematically meaningful, challenging question that fits the label path.
 		
 		\textbf{\#\# Instructions}
 		
 		\textbf{\#\#\# Step 1: Path-Guided Complex Information Extraction}\\
 		\textbf{**PRIORITY: Use the label path to focus on mathematically relevant, non-trivial content.**}
 		
 		\begin{enumerate}[leftmargin=1.5em,itemsep=0.15em,topsep=0.1em,parsep=0pt,partopsep=0pt]
 			\item Interpret the label path:
 			\begin{itemize}[leftmargin=1.4em,itemsep=0.1em,topsep=0.1em,parsep=0pt,partopsep=0pt]
 				\item Identify the main domain (e.g., \texttt{"Mathematics"}).
 				\item Identify intermediate sub-domains (e.g., \texttt{"Representation Theory"}, \texttt{"Lie Theory"}).
 				\item Identify the atomic knowledge point (the last label).
 			\end{itemize}
 			
 			\item Scan the text and identify information that:
 			\begin{itemize}[leftmargin=1.4em,itemsep=0.1em,topsep=0.1em,parsep=0pt,partopsep=0pt]
 				\item Is directly about the atomic knowledge point, or
 				\item Naturally belongs to the specified path (e.g., theorems, constructions, examples, or techniques in that subfield).
 			\end{itemize}
 			
 			\item Among that information, focus on content that requires connecting multiple ideas, such as:
 			\begin{itemize}[leftmargin=1.4em,itemsep=0.1em,topsep=0.1em,parsep=0pt,partopsep=0pt]
 				\item Relationships between several mathematical objects (groups, modules, functors, root systems, etc.).
 				\item Multi-step derivations, proofs, or constructions.
 				\item Interactions between definitions, lemmas, and theorems.
 				\item Situations where properties at a higher level in the path (e.g., \texttt{"Representation Theory"}) constrain or influence the atomic concept.
 			\end{itemize}
 		\end{enumerate}
 		
 		\textbf{**AVOID**:}
 		\begin{itemize}[leftmargin=1.4em,itemsep=0.1em,topsep=0.1em,parsep=0pt,partopsep=0pt]
 			\item Generic reasoning or non-mathematical content, even if it appears in the text.
 			\item Simple, standalone definitions that require no reasoning.
 			\item Single, directly stated facts that can be copied as-is.
 			\item Questions that do not clearly live inside the given label path.
 		\end{itemize}
 		
 		Your goal is to pick a relationship or conclusion that:
 		\begin{itemize}[leftmargin=1.4em,itemsep=0.1em,topsep=0.1em,parsep=0pt,partopsep=0pt]
 			\item Is genuinely about the atomic knowledge point AND its mathematical context.
 			\item Requires synthesis of multiple pieces of mathematical information.
 		\end{itemize}
 		
 		\textbf{\#\#\# Step 2: Difficulty Enhancement Process}\\
 		\textbf{**EXPLICITLY STATE YOUR HARDENING PROCESS**}
 		
 		Before generating the question, describe your strategy to make it harder:
 		\begin{enumerate}[leftmargin=1.5em,itemsep=0.1em,topsep=0.1em,parsep=0pt,partopsep=0pt]
 			\item What simple version would you avoid?
 			\item What complexity layers will you add?
 			\item Which concepts will you force students to connect?
 			\item What common shortcuts will you block?
 			\item How will you ensure multi-step reasoning is required?
 		\end{enumerate}
 		
 		Document this in the output field \texttt{"hardening\_process"}.
 		
 		\textbf{\#\#\# Step 3: Advanced Question Generation}
 		
 		For each complex relationship identified, create a question that:
 		\begin{itemize}[leftmargin=1.4em,itemsep=0.1em,topsep=0.1em,parsep=0pt,partopsep=0pt]
 			\item Requires applying multiple concepts from different parts of the document
 			\item Tests understanding of relationships, not just recall of facts
 			\item Forces reasoning through multiple steps to reach the answer
 			\item May require comparing or contrasting different scenarios
 			\item Could involve \texttt{"what if"} scenarios based on principles in the text
 			\item Tests ability to apply concepts to slightly modified situations
 		\end{itemize}
 		
 	\end{tcolorbox}
 \end{figure*}

   \begin{figure*}[t]
   	\vspace{-20mm}
	\centering
	\begin{tcolorbox}[
		enhanced,
		breakable,
		colback=white,
		colframe=blue!50!red,
		colbacktitle=blue!50!red,
		coltitle=white,
		title={MCQ Question Generation Prompt (Path-Conditioned, adapted from SPICE) (Continued)},
		center title,
		fonttitle=\bfseries,
		boxrule=1.2pt,
		arc=1.8mm,
		left=2.2mm,
		right=1.2mm,
		top=0.2mm,
		bottom=0.2mm,
		boxsep=0.2mm
		]
		\footnotesize
		
		\textbf{**CRITICAL - Self-Contained Requirements**:}
		\begin{itemize}[leftmargin=1.4em,itemsep=0.1em,topsep=0.1em,parsep=0pt,partopsep=0pt]
			\item Questions must be 100\% self-contained and standalone
			\item NEVER use phrases like: \texttt{"according to the text"}, \texttt{"in the document"}, \texttt{"as mentioned"}, \texttt{"the passage states"}, \texttt{"based on the analysis"}, etc.
			\item Write as if for a formal exam with no reference material
			\item Include all necessary context within the question itself
			\item Define any specialized terms if needed for clarity
		\end{itemize}
		
		\textbf{\#\#\# Step 4: Difficulty-Driven Design}\\
		\textbf{**TARGET: Generate HARD/EXTRA HARD questions by design**}
		\begin{itemize}[leftmargin=1.4em,itemsep=0.1em,topsep=0.1em,parsep=0pt,partopsep=0pt]
			\item HARD: Synthesize 4+ concepts; multi-step problem solving; pattern recognition
			\item EXTRA HARD: Complex system analysis; counter-intuitive applications; edge cases
		\end{itemize}
		
		Design questions that \textbf{CANNOT} be answered by:
		\begin{itemize}[leftmargin=1.4em,itemsep=0.1em,topsep=0.1em,parsep=0pt,partopsep=0pt]
			\item Looking up a single fact
			\item Finding one sentence with the answer
			\item Simple keyword matching
		\end{itemize}
		
		\textbf{\#\#\# Step 5: Knowledge Integration Requirements}
		
		Document the reasoning path that shows why this is a difficult question:
		\begin{itemize}[leftmargin=1.4em,itemsep=0.1em,topsep=0.1em,parsep=0pt,partopsep=0pt]
			\item List 3+ distinct pieces of information needed from different parts of the document
			\item Show the logical connections required between these pieces
			\item Explain why simple lookup won't work
			\item Include intermediate reasoning steps
		\end{itemize}
		
		\textbf{\#\#\# Step 6: Multiple Choice Design Guidelines}
		
		Create a multiple choice question with 4 options following these STRICT rules:
		\begin{itemize}[leftmargin=1.4em,itemsep=0.1em,topsep=0.1em,parsep=0pt,partopsep=0pt]
			\item \textbf{Length Balance}: All options must be approximately equal length (+/- 20\%).
			\item \textbf{Unit Consistency}: All numerical answers must use identical units and formatting.
			\item \textbf{Tone Neutrality}: Avoid overly certain language (\texttt{"definitely"}, \texttt{"always"}, \texttt{"never"}) unless justified.
			\item \textbf{Plausibility}: All distractors must be genuinely plausible based on partial understanding.
		\end{itemize}
		
		Format:\\
		Question: [Complete, self-contained question with all necessary context]\\
		A) [Balanced length option]\\
		B) [Balanced length option]\\
		C) [Balanced length option]\\
		D) [Balanced length option]
		
		\textbf{**Distractor Design**:}
		\begin{itemize}[leftmargin=1.4em,itemsep=0.1em,topsep=0.1em,parsep=0pt,partopsep=0pt]
			\item Common calculation errors from the multi-step process
			\item Results from applying only partial reasoning
			\item Mixing up related concepts from the document
			\item Reasonable approximations that miss key factors
		\end{itemize}
		
		\textbf{\#\#\# Step 7: Self-Testing Filter (AFTER MCQ Creation)}\\
		\textbf{**SOLVE YOUR OWN MCQ AS A STUDENT WOULD**}
		
		Now test the complete multiple choice question:
		\begin{enumerate}[leftmargin=1.5em,itemsep=0.1em,topsep=0.1em,parsep=0pt,partopsep=0pt]
			\item What's the quickest path a student might try with these options?
			\item Can you eliminate 2+ options without full understanding? If yes, redesign distractors.
			\item Does seeing the options make the answer obvious? If yes, improve distractors.
			\item Count the reasoning steps required even with options visible - if less than 3, REJECT.
			\item Time estimate: Would this MCQ take <30 seconds? If yes, make it harder.
			\item Could a student guess correctly by pattern matching the options? If yes, rebalance.
		\end{enumerate}
		
		Document your solving process in \texttt{"self\_test\_solution"}.
		
		\textbf{\#\#\# Step 8: Final Complexity Verification}
		
		Before finalizing, verify your question is \textbf{NOT} Easy by checking:
		\begin{itemize}[leftmargin=1.4em,itemsep=0.1em,topsep=0.1em,parsep=0pt,partopsep=0pt]
			\item Can it be answered by finding one sentence? If yes, redesign.
			\item Does it require connecting multiple document sections? If no, add complexity.
			\item Would someone need to understand relationships, not just facts? If no, refocus.
			\item Are all MCQ options balanced and using consistent formatting? If no, revise.
			\item Did your self-test of the MCQ take more than 1 minute? If no, increase difficulty.
		\end{itemize}
		
		\textbf{\#\# Output Format}
		
		FIRST, think step-by-step about your question design (this is your private thinking).\\
		THEN, provide your complete analysis in a JSON object with these fields.
		
		CRITICAL: Output \textbf{ONLY} valid JSON without any markdown formatting or code blocks.\\
		DO \textbf{NOT} wrap your JSON in \texttt{```} or \texttt{"json"} markers.\\
		Start directly with \texttt{\{} and end with \texttt{\}}.
		
		Required fields:
		\begin{itemize}[leftmargin=1.4em,itemsep=0.1em,topsep=0.1em,parsep=0pt,partopsep=0pt]
			\item \texttt{"identified\_answer"}
			\item \texttt{"answer\_quote"}
			\item \texttt{"hardening\_process"}
			\item \texttt{"question"}
			\item \texttt{"correct\_answer"}
			\item \texttt{"self\_test\_solution"}
			\item \texttt{"knowledge\_and\_reasoning\_steps"}
			\item \texttt{"question\_difficulty"}
		\end{itemize}
		
	\end{tcolorbox}
\end{figure*}

    \begin{figure*}[t]
 	\centering
 	\begin{tcolorbox}[
 		enhanced,
 		breakable,
 		colback=white,
 		colframe=blue!50!red,
 		colbacktitle=blue!50!red,
 		coltitle=white,
 		title={Free-form Question Generation Prompt (Path-Conditioned, adapted from SPICE)},
 		center title,
 		fonttitle=\bfseries,
 		boxrule=1.2pt,
 		arc=1.8mm,
 		left=1.2mm,
 		right=1.2mm,
 		top=0.4mm,
 		bottom=0.4mm,
 		boxsep=0.4mm
 		]
 		\footnotesize
 		
 		Your task is to create a \textbf{CHALLENGING} question from a document by using \textbf{BOTH}:
 		\begin{itemize}[leftmargin=1.4em,itemsep=0.1em,topsep=0.1em,parsep=0pt,partopsep=0pt]
 			\item (1) a hierarchical \textbf{LABEL PATH} that narrows down the the mathematical domain, and
 			\item (2) background \textbf{TEXT} about the most specific knowledge point.
 		\end{itemize}
 		
 		\textbf{\#\# Label Path (Domain Hierarchy)}\\
 		\texttt{[BEGINNING OF THE LABEL PATH]}\\
 		\texttt{\{path\}}\\
 		\texttt{[END OF THE LABEL PATH]}
 		
 		The label path lists nested mathematical domains from the broadest on the left to the most specific on the right.\\
 		Example: \texttt{"Mathematics -> Algebra -> Group Theory -> Sylow's Theorems"}
 		
 		\begin{itemize}[leftmargin=1.4em,itemsep=0.1em,topsep=0.1em,parsep=0pt,partopsep=0pt]
 			\item The \textbf{LEFTMOST} labels are broad fields (e.g., \texttt{"Mathematics"}, \texttt{"Algebra"}).
 			\item The \textbf{RIGHTMOST} label is an \textbf{ATOMIC KNOWLEDGE POINT} (e.g., \texttt{"Sylow's Theorems"}).
 			\item Your question \textbf{MUST} belong to this path:
 			\begin{itemize}[leftmargin=1.4em,itemsep=0.1em,topsep=0.1em,parsep=0pt,partopsep=0pt]
 				\item It must clearly be a mathematics question.
 				\item It must primarily test the \textbf{RIGHTMOST} knowledge point.
 				\item It may use context from earlier levels in the path to add difficulty and require multi-step reasoning.
 			\end{itemize}
 			\item If the text contains information that is irrelevant to this label path, \textbf{IGNORE} that information.
 		\end{itemize}
 		
 		\textbf{\#\# Text}\\
 		\texttt{[BEGINNING OF THE DOCUMENT]}\\
 		\texttt{\{text\}}\\
 		\texttt{[END OF THE DOCUMENT]}
 		
 		The text is background material (e.g., web pages) about the atomic knowledge point at the end of the label path.\\
 		You must use this text to construct a mathematically meaningful, challenging question that fits the label path.
 		
 		\textbf{\#\# Instructions}
 		
 		\textbf{\#\#\# Step 1: Path-Guided Complex Information Extraction}\\
 		\textbf{**PRIORITY: Use the label path to focus on mathematically relevant, non-trivial content.**}\\
 		\texttt{... (same as above) ...}
 		
 		\textbf{\#\#\# Step 3: Advanced Question Generation (Free-Form Answer)}
 		
 		For each complex relationship identified, create a question that:
 		\begin{itemize}[leftmargin=1.4em,itemsep=0.1em,topsep=0.1em,parsep=0pt,partopsep=0pt]
 			\item Requires applying multiple concepts from different parts of the document
 			\item Tests understanding of relationships, not just recall of facts
 			\item Forces reasoning through multiple steps to reach the answer
 			\item May require comparing or contrasting different scenarios
 		\end{itemize}
 		
 		The answer must be a typed free-form answer extracted or computed from the document:
 		\begin{itemize}[leftmargin=1.4em,itemsep=0.1em,topsep=0.1em,parsep=0pt,partopsep=0pt]
 			\item A numeric value (integer or real number)
 			\item A symbolic or algebraic expression
 			\item A short string (name, label, or concept) that appears in or is uniquely determined by the document
 		\end{itemize}
 		
 		\textbf{\#\#\# Step 4--8}\\
 		\texttt{(Identical self-contained requirements, difficulty checks, and self-testing as in the MCQ prompt,}\\
 		\texttt{except that the output is a single free-form answer rather than an option letter.)}
 		
 		\textbf{\#\# Output Format}
 		
 		Output \textbf{ONLY} valid JSON with required fields:
 		\begin{itemize}[leftmargin=1.4em,itemsep=0.1em,topsep=0.1em,parsep=0pt,partopsep=0pt]
 			\item \texttt{"identified\_answer"}
 			\item \texttt{"answer\_quote"}
 			\item \texttt{"hardening\_process"}
 			\item \texttt{"question"}
 			\item \texttt{"correct\_answer"}
 			\item \texttt{"answer\_type"}
 			\item \texttt{"self\_test\_solution"}
 			\item \texttt{"knowledge\_and\_reasoning\_steps"}
 			\item \texttt{"question\_difficulty"}
 		\end{itemize}
 		
 	\end{tcolorbox}
 \end{figure*}

\end{document}